\pdfoutput=1
\documentclass[11pt]{article}
\usepackage[preprint]{acl}

\usepackage[utf8]{inputenc}
\usepackage[T1]{fontenc}
\usepackage{times}
\usepackage{latexsym}
\usepackage{booktabs}
\usepackage{amsmath}
\usepackage{amssymb}
\usepackage{amsthm}
\usepackage{graphicx}
\usepackage{xcolor}
\usepackage{microtype}
\usepackage{nicefrac}
\usepackage{multirow}
\usepackage{diagbox}
\usepackage{algorithm}
\usepackage{algpseudocode}
\usepackage{subcaption}
\usepackage{xspace}
\usepackage{inconsolata}
\usepackage{CJKutf8}

\newcommand{\rkeep}{r_{\mathrm{keep}}}

\usepackage{placeins}

\title{Text-Preserving Lossy Text Compression:
       A Study of Strategic Deletion and LLM Reconstruction}

\author{Yuchun Zou \\
  CUNY Graduate Center \\
  \texttt{yzou@gc.cuny.edu} \And
  Junhong Tong \\
  CUNY Queens College \\
  \texttt{junhong.tong16@} \\
  \texttt{stu-mail.qc.cuny.edu} \And
  Jun Li \\
  CUNY Queens College \\
  \& Graduate Center \\
  \texttt{jun.li@qc.cuny.edu}}

\begin{document}
\maketitle

\begin{abstract}
Traditional lossless text compression preserves every byte, but its gains on natural language are often modest. We study \emph{lossy semantic text compression}, where the encoder strategically deletes parts of the text and a large language model (LLM) reconstructs the original from the retained skeleton. We benchmark a progression of deletion strategies, from character-level and word-level rules driven by word length or corpus frequency to surprisal-based and hybrid rules that require a neural encoder, and reach three findings. First, deletion by corpus word frequency is a strong low-cost baseline: a static frequency lookup remains competitive with far more expensive semantic methods. Second, semantic and hybrid methods gain most at mild-to-moderate compression, whereas at the lowest retention rate frequency-based deletion matches them and keeps more named entities. Third, QLoRA fine-tuning yields a compact local decoder that, once adapted, is competitive with zero-shot Gemini 2.0 Flash.\footnote{Gemini~2.0 Flash is the decoder for every result in this paper unless a table states Gemini~2.5 Flash. Google retired 2.0 Flash from its API during the review period, so experiments added after submission use its successor, 2.5 Flash. Re-decoding the WordFreq-3 skeletons with 2.5 Flash reproduced every aggregate BERTScore within $0.016$~F1 (Table~\ref{tab:decoder_check}).} Additional English and Chinese experiments show that the framework transfers across domains, while the best deletion rule remains dataset-dependent.
\end{abstract}

\section{Introduction}
\label{sec:intro}

\begin{figure*}[t]
\centering
\includegraphics[width=\textwidth]{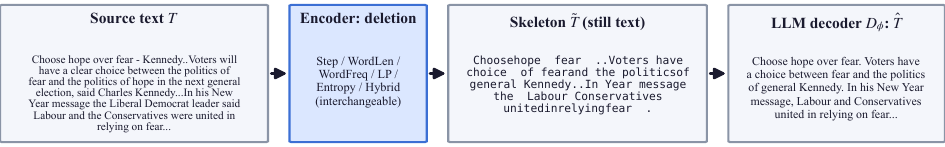}
\caption{The pipeline on a verbatim BBC example (WordFreq-6, $\rkeep = 0.5$): the encoder deletes tokens, the
skeleton stays text and can be searched, edited or losslessly compressed before any decoder runs,
and the LLM decoder reconstructs from the skeleton alone. The reconstruction keeps a genuine distortion
(\emph{general election} $\rightarrow$ \emph{general Kennedy}).}
\label{fig:pipeline}
\end{figure*}

The information capacity of digital networks and storage systems has grown enormously, yet the volume of generated text data continues to outpace these expansions. Traditional lossless compression algorithms, such as zlib~\citep{zlib}, bzip2~\citep{bz2}, and LZMA~\citep{lzma}, reduce storage requirements by exploiting statistical redundancies at the byte level. However, because they guarantee exact, bit-for-bit reconstruction of the original sequence, their compression gains on natural language are inherently limited and often remain modest in realistic operating regimes.

\textbf{Motivation.} Our primary target setting is context- and bandwidth-limited processing of factual text in LLM-centered pipelines. A document or prior interaction history may need to pass through a narrow text channel before later use. For example, it may be inserted into a limited context window, cached as agent memory, relayed between text-based components, or uploaded from a weak sender to a stronger receiver. These settings share a cost asymmetry: the encoder runs once on every chunk written, while the decoder runs only on the chunks whose full text is later requested, a small fraction in logs, archives, and agent memory. In such settings, the intermediate representation may still need to remain textual rather than become an opaque latent vector: a textual skeleton passes through existing text interfaces (storage, logging, transmission, version control) unchanged, and when whole words are kept it remains searchable and even usable without any decoder (Appendix~\ref{app:downstream}). This makes \emph{text-preserving} lossy compression a distinct operating regime from unconstrained semantic coding. We do not claim that a textual bottleneck is information-theoretically optimal; rather, we study the practical regime where the compressed artifact must remain text and approximate semantic recovery is acceptable. Yet, to our knowledge, no systematic benchmark exists for this regime, leaving practitioners without principled guidance on deletion strategy, encoder-side cost, or the viability of lightweight local decoders.

We address this gap in this paper. Our framework (Figure~\ref{fig:pipeline}) transmits a compressed skeleton obtained by strategically deleting characters or tokens, and delegates reconstruction to an LLM decoder at the receiver. This directly targets settings such as context-window reduction, agent-memory caching, and weak-sender/strong-receiver text transfer, where the artifact being compressed is meant to remain useful as text before or after reconstruction. Unlike abstractive summarization~\citep{rush2015neural,see2017get,liu2019text}, which rewrites the source, our approach preserves original tokens as anchor points for reconstruction. Unlike prompt compression methods such as LLMLingua~\citep{jiang2023llmlingua} and Selective Context~\citep{li2023compressing}, which optimize downstream task accuracy for a fixed consumer, our goal is to maximize fidelity to the \emph{original text} after reconstruction. We therefore study reusable lossy text artifacts for later recovery, not only immediate task performance from a compressed prompt. Appendix~\ref{app:related_work} surveys related work.

Determining \emph{what} to delete is critical: na\"ive uniform deletion destroys linguistic context and maximizes ambiguity. We compare a progression of deletion schemes: uniform step deletion, word-length-guided deletion (WordLen), word-frequency-guided deletion (WordFreq), deletion guided by an allocation of the budget across frequency buckets (LP), entropy-based deletion via GPT-2 surprisal, and hybrid combinations of frequency and surprisal signals.
We evaluate via post-reconstruction BERTScore~\citep{zhang2020bertscore} on English news across retention rates $\rkeep \in [0.1, 0.9]$, using zero-shot Gemini 2.0 Flash and QLoRA-fine-tuned Llama-3.2-3B-Instruct~\citep{dettmers2023qlora} as decoders. Cross-domain experiments on additional English datasets (Wikipedia, Reddit) and an extension to Chinese news are reported in the appendix. We also include a length-constrained LLM summarization baseline to separate reconstruction-oriented deletion from a more conventional lossy rewriting alternative.

\textbf{Contributions.} \textit{(i)} We formulate reconstruction-oriented, text-preserving lossy compression as a benchmark setting with a systematic evaluation protocol (BERTScore, ROUGE-L, and entity recall on the same reconstructions, with human and LLM fact-level judgments, further lexical metrics, and downstream tasks in the appendices) across nine retention rates. \textit{(ii)} We provide a measurement study over a broad family of deletion strategies, showing that their relative strengths are strongly regime-dependent: semantic and hybrid methods help most at mild-to-moderate compression, while WordFreq matches them at the lowest retention rate and preserves more named entities there. \textit{(iii)} We identify low-resource sender-side compression schemes that are practical when encoder compute, memory, power, or bandwidth are constrained: in particular, WordFreq achieves competitive reconstruction quality with only a static lookup, while surprisal-based hybrids buy a small further gain at mild compression at a far higher encoder cost. \textit{(iv)} We show that strategy-aware fine-tuning can make a compact local decoder highly competitive with a stronger zero-shot proprietary decoder under the same reconstruction setting, enabling a local deployment path. \textit{(v)} Cross-domain experiments on four datasets (see Appendices~\ref{app:wikipedia_eval}--\ref{app:chinese_eval}) show that several qualitative trends transfer, but the best deletion rule remains domain-dependent.

\section{Problem Formulation}
\label{sec:problem}

We formulate lossy semantic text compression as a two-phase encode-decode problem (Figure~\ref{fig:pipeline}). Given a source text $T$ of length $L$ and a target retention rate $\rkeep \in (0,1)$, the encoder produces a degraded representation $\tilde{T}$ ({\em i.e.}, the skeleton) of length $\rkeep \cdot L$ by strategically deleting components from $T$, and the decoder $D_\phi$ reconstructs $\hat{T} = D_\phi(\tilde{T})$ from the degraded input alone. Because the
skeleton is still text, it can be compressed further by a lossless codec and decoded exactly before
reconstruction, leaving the reconstructed text unchanged. We therefore report storage in three tiers, of which $1/\rkeep$ is only the first
(Appendix~\ref{app:lossless_baselines}), since $1/\rkeep$ alone is not what a deployment stores. At inference time, only $\tilde{T}$ and lightweight strategy metadata need to be stored or transmitted; decoder weights and prompt templates are treated as shared side information. This setup is intentionally narrower than continuous latent semantic coding: our focus is on compression schemes whose intermediate representation remains text.

We evaluate reconstruction quality primarily with BERTScore~\citep{zhang2020bertscore}, averaged over reconstructed chunks, and report ROUGE-L~\citep{lin2004rouge} and Character Error Rate (CER) as complementary lexical-fidelity metrics. BERTScore and ROUGE-L appear in the main text. CER, chrF++ and METEOR are reported in Appendix~\ref{app:detailed_metrics}, which also tests whether the choice of metric changes which strategy wins. Because no single automatic metric fully captures factual faithfulness, we also report entity recall and fact-level judgments (Appendices~\ref{app:human_eval}--\ref{app:fact_eval}).

\section{Methodology}
\label{sec:method}

This section describes the full encode--decode pipeline. On the encoder side, \emph{strategic deletion} compresses a source text into a textual skeleton under a retention budget. On the decoder side, \emph{LLM-based semantic reconstruction} expands that skeleton into a fuller version of the source.

\subsection*{Encoder: Structured Degradation Strategies}

We organize the encoder strategies into three levels of linguistic sophistication: \textbf{Level~1} treats text as a character stream, \textbf{Level~2} respects word boundaries and uses corpus-frequency statistics, and \textbf{Level~3} uses a neural language model to estimate contextual predictability. Higher levels improve token selection at higher encoder cost, a trade-off quantified in Section~\ref{sec:new_baselines}.

\subsection*{Level 1: Character-Level Deletion}

\label{sec:step}
We first establish a simple baseline: \textit{fixed-step character deletion}. To achieve a target retention rate $\rkeep$, this strategy deletes every $\lfloor 1/(1-\rkeep) \rfloor$-th character, spaces and punctuation included, producing a uniformly subsampled sequence.

While straightforward, this uniform deletion strategy is entirely agnostic to linguistic structure. It treats all characters identically ({\em i.e.}, letters, digits, whitespace, and punctuation alike), blindly destroying critical morphological boundaries and context. Consequently, it serves as a strict lower-bound benchmark, motivating the need for structure-aware compression. We also evaluate three stochastic character-deletion baselines (Gaussian, Bernoulli, and Poisson) that draw deletion positions from the named distribution.

\subsection*{Level 2: Word-Level Deletion}

\textbf{Adaptive Small-Word Removal (WordLen).}
\label{sec:wordlen}
To overcome the destructive nature of uniform character deletion, we propose an \textit{adaptive small-word removal} strategy. Instead of blind subsampling, it progressively applies structured edits to remove components with lower semantic importance, aiming to reach a tolerance interval $[\rkeep - \epsilon, \rkeep]$.

The algorithm proceeds through increasingly aggressive monotonic transformations: whitespace reduction; vowel deletion in words of length $\ge 3$ ({\em e.g.}, \texttt{documentation} $\to$ \texttt{dcmnttn}); short-word deletion or truncation for 1--2 character words; long-word shortening that preserves initial stems; punctuation and number removal; and, if needed, random character deletion from progressively shorter words. This method preserves stems and structural anchors much better than the fixed-step benchmark, but length is an imperfect proxy for semantic importance and the multi-stage pipeline is difficult to tune. We therefore treat it as a conceptual stepping stone and focus the main analysis on the frequency-based methods below, including the LP variant.

\textbf{Frequency-Based Static Deletion (WordFreq).}
\label{sec:wordfreq}
This strategy uses Zipf frequency scores~\citep{speer2022wordfreq} to prioritize the removal of highly predictable, high-frequency words while preserving rare, information-dense ones.

Tokens are mapped to three broad frequency classes: \textsc{low} (Zipf $\le 3$), \textsc{mid} ($3 <$ Zipf $\le 5$), and \textsc{high} (Zipf $> 5$). The strategy computes the natural distribution of these classes in the source text. When a target number of characters must be deleted to meet the budget, the deletion quota is distributed proportionally according to the original class frequencies. Within each class, whole words are then removed uniformly at random. Punctuation and numerals are kept.
Appendix~\ref{app:ablations} compares this three-class taxonomy with a six-class one and reports why we keep three as the default. 

This static approach grounds compression in semantic redundancy. However, the fixed proportional allocation forces arbitrary destruction across all classes, and the coarse three-bucket grouping may not fully capture the distinct roles of punctuation, whitespace, and numbers.

\textbf{Frequency-Aware LP-Guided Deletion (LP).}
\label{sec:optimize}
To achieve finer control over information removal, we extend the static frequency model with a bucket-level allocation of the deletion budget. We first expand the token taxonomy to six buckets: \textsc{low}, \textsc{mid}, \textsc{high} (the same Zipf classes), \textsc{punct}, \textsc{others}, and \textsc{whitespace}. Rather than using strictly proportional deletion, we choose class-specific deletion ratios $w_k \in [0,1]$ that maximize a first-order estimate of semantic preservation.

Let $p_k$ denote the proportion of characters in bucket $k$. We model the BERTScore retained when a fraction $w_k$ of bucket $k$ is deleted as a linear factor
$
B_k(w_k) = 1 - w_k(1 - B_k^{\mathrm{full}})
$,
where $B_k^{\mathrm{full}}$ is the BERTScore of the decoded output when the entire bucket is deleted, and treat the buckets as independent, so the estimated score after deletion is the product of the factors. The ratios are chosen by solving
\begin{align*}
  \max_{\{w_k\}} \quad & \prod_k B_k(w_k) \\
  \text{s.t.} \quad & \sum_k p_k w_k \ge 1 - \rkeep, \notag \\
                    & 0 \le w_k \le 1 \quad \forall k. \notag
\end{align*}
We use the name LP because the first-order expansion of this objective is a linear program (Appendix~\ref{app:lp_derivation}). The solution sacrifices the token classes that cost the least score per unit of deletion budget before touching highly sensitive classes. LP-guided deletion is a first-order heuristic: the allocation is the best one for this bucket-level estimate, not for reconstruction quality itself. The estimate is intentionally coarse, to keep the allocation interpretable and tractable. It treats each bucket's contribution as linear in its deletion ratio, between no deletion and full deletion of the bucket, and it treats buckets as independent, with no interaction terms. It is therefore least reliable when the budget forces several buckets to be emptied at once. Section~\ref{sec:main_results} reports the regimes where LP leads and where it falls behind, and Appendix~\ref{app:lp_derivation} gives the derivation.

\subsection*{Level 3: Semantic-Level Deletion}

\textbf{Entropy-Based Deletion.}
\label{sec:entropy}
Frequency is only a static proxy for predictability. A more precise signal is per-token surprisal under a language model, which measures contextual information content directly. Following the general principle of LLMLingua~\citep{jiang2023llmlingua}, we compute GPT-2~\citep{radford2019language} surprisal and delete the lowest-surprisal tokens first. This improves token selection, especially at stronger compression, but requires neural encoder inference (Table~\ref{tab:latency}).

\textbf{Hybrid Frequency--Entropy Deletion.}
\label{sec:hybrid_method}
Frequency and surprisal capture complementary signals: global corpus redundancy versus contextual predictability. We therefore study three hybrids. \textbf{Entropy-LP} replaces the Zipf buckets in the LP allocation with surprisal tertiles, changing \emph{how much} budget each group receives. \textbf{Entropy-in-FreqBuckets} keeps the frequency-based LP allocation but deletes lowest-surprisal tokens within each frequency bucket, changing \emph{which} tokens are removed. \textbf{Hybrid-$\alpha$} combines frequency rank and surprisal rank into a single token-level score, with $\alpha \in \{0.3,0.5,0.7\}$ controlling the interpolation between WordFreq-like and entropy-like behavior.

\textbf{LLM Summarization Baseline.}
\label{sec:summarization}
As a contrasting paradigm, Gemini is prompted to rewrite the original text to $\rkeep \times |\text{original}|$ characters while preserving key facts~\citep{see2017get}. There is no skeleton and nothing to reconstruct.

\subsection*{Decoder: Semantic Reconstruction}

\textbf{LLM-Based Semantic Reconstruction.}
\label{sec:llm}
Following compression, a generative LLM acts as the decoder $D_\phi(\tilde{T})$. We utilize Gemini 2.0 Flash~\citep{geminiteam2025gemini20flash} as the primary zero-shot decoder. The model is prompted with the skeleton $\tilde{T}$ and instructed to reconstruct fluent text that preserves the original meaning without adding new facts. To curb output divergence ({\em i.e.}, models either truncating difficult passages or inventing content), we enforce a length constraint ($\pm 15\%$ of estimated original length), applying retry logic if violated.

\textbf{Strategy-Aware Fine-Tuning.}
\label{sec:finetune}
As a compute-efficient alternative to proprietary zero-shot LLMs for reconstruction, we introduce \textit{strategy-aware supervised fine-tuning} (SFT) of the decoder using Llama-3.2-3B-Instruct~\citep{dubey2024llama3} and QLoRA~\citep{dettmers2023qlora}. The model is trained on a mixture of ratio-specific degradation patterns, learning the causal language modeling objective to predict the original text $T$ given the compressed prefix $\tilde{T}$. By exposing the model directly to the artifacts of our structured deletion schemes, it internalizes compression-aware reconstruction capabilities, allowing a lightweight 3B model to compete with massive closed-source counterparts.

\begin{table*}[t]
\centering
\small
\setlength{\tabcolsep}{5pt}
\caption{BERTScore F1 for character-level deletion methods on $100$ BBC chunks, decoded with Gemini~2.5 Flash~\citep{geminiteam2025gemini25}.
Std.\ deviations are $0.004$--$0.020$.
\textbf{Bold} = best per column.}
\label{tab:char_deletion}
\begin{tabular}{lccccccccc}
\toprule
\diagbox{\textbf{Method}}{$\rkeep$} & \textbf{0.9} & \textbf{0.8} & \textbf{0.7}
  & \textbf{0.6} & \textbf{0.5} & \textbf{0.4}
  & \textbf{0.3} & \textbf{0.2} & \textbf{0.1} \\
\midrule
Gaussian  & 0.9823 & 0.9706 & 0.9582 & \textbf{0.9431} & \textbf{0.9307} & \textbf{0.9174} & \textbf{0.9043} & \textbf{0.8911} & \textbf{0.8759} \\
Bernoulli & 0.9900 & 0.9826 & 0.9637 & 0.9295 & 0.8991 & 0.8763 & 0.8654 & 0.8596 & 0.8556 \\
Poisson   & 0.9913 & 0.9847 & 0.9644 & 0.9233 & 0.8854 & 0.8668 & 0.8607 & 0.8570 & 0.8539 \\
\textbf{Step (Uniform)} & \textbf{0.9916} & \textbf{0.9880} & \textbf{0.9708} & 0.9129 & 0.8701 & 0.8652 & 0.8604 & 0.8586 & 0.8551 \\
\bottomrule
\end{tabular}
\end{table*}

\begin{table*}[t]
\centering
\small
\setlength{\tabcolsep}{6pt}
\caption{Word-level deletion strategies under three lenses applied to the \emph{same} reconstructions:
semantic (BERTScore F1), lexical (ROUGE-L F1), and factual (entity recall, the share of the original's named-entity mentions present in the reconstruction; $n{=}200$--$201$ articles, $8{,}021$ mentions).
\textbf{Bold} = best per column within a block. Standard deviations, the Step baseline, character error
rate, and the unsupported-entity rate that accompanies entity recall are in
Appendix~\ref{app:detailed_metrics} and Table~\ref{tab:ner_recon}.}
\label{tab:unified}
\begin{tabular}{lccccccccc}
\toprule
\diagbox{\textbf{Method}}{$\rkeep$} & \textbf{0.9} & \textbf{0.8} & \textbf{0.7} & \textbf{0.6} & \textbf{0.5} & \textbf{0.4} & \textbf{0.3} & \textbf{0.2} & \textbf{0.1} \\
\midrule
\multicolumn{10}{l}{\emph{BERTScore F1: semantic similarity}} \\
WordLen  & 0.9884 & \textbf{0.9885} & \textbf{0.9865} & \textbf{0.9827} & 0.9475 & 0.9176 & 0.8485 & 0.8220 & 0.8096 \\
WordFreq & 0.9839 & 0.9754 & 0.9645 & 0.9479 & 0.9314 & 0.9137 & \textbf{0.8948} & \textbf{0.8737} & \textbf{0.8527} \\
LP      & \textbf{0.9885} & 0.9862 & 0.9780 & 0.9668 & \textbf{0.9484} & \textbf{0.9211} & 0.8924 & 0.8665 & 0.8397 \\
\addlinespace[3pt]
\multicolumn{10}{l}{\emph{ROUGE-L F1: lexical overlap}} \\
WordLen  & 0.950 & \textbf{0.953} & \textbf{0.942} & \textbf{0.918} & 0.714 & 0.588 & 0.210 & 0.111 & 0.073 \\
WordFreq & 0.923 & 0.878 & 0.824 & 0.752 & 0.664 & 0.568 & 0.449 & \textbf{0.323} & \textbf{0.191} \\
LP      & \textbf{0.954} & 0.940 & 0.889 & 0.826 & \textbf{0.748} & \textbf{0.628} & \textbf{0.478} & 0.312 & 0.142 \\
\addlinespace[3pt]
\multicolumn{10}{l}{\emph{Entity recall: factual preservation}} \\
WordLen  & \textbf{0.978} & \textbf{0.974} & \textbf{0.947} & \textbf{0.914} & \textbf{0.760} & 0.624 & 0.264 & 0.102 & 0.023 \\
WordFreq & 0.937 & 0.894 & 0.854 & 0.798 & 0.722 & \textbf{0.647} & \textbf{0.544} & \textbf{0.417} & \textbf{0.268} \\
LP      & 0.954 & 0.935 & 0.883 & 0.800 & 0.653 & 0.529 & 0.398 & 0.285 & 0.191 \\
\bottomrule
\end{tabular}
\end{table*}

\section{Experiments}

\label{sec:experiments}
\subsection{Experimental Setup}
\textbf{Dataset.}
\label{sec:dataset}
Our main benchmark is the BBC News dataset~\citep{kalpande2023bbcnews}, a collection of newswire
articles covering domestic and international news.  We split articles into chunks of at most $512$ characters.  For evaluation we use a test set of $200$ held-out article chunks. For fine-tuning, we use a dataset of $2{,}000$ training chunks disjoint from the evaluation set. To prevent overfitting and ensure generalization, we reserve $10\%$ of this training data as a validation set, monitoring the validation loss during training and selecting the optimal checkpoint once the loss stabilizes. All main-paper experiments use this dataset. Cross-domain results on Wikipedia (Appendix~\ref{app:wikipedia_eval}), Reddit (Appendix~\ref{app:reddit_eval}), and Chinese text (Appendix~\ref{app:chinese_eval}) are in the appendices.

\textbf{Tokenization.}
All English texts are tokenized using standard word boundaries,
producing an average of 5.0 characters per word token.
The six frequency buckets are populated using Zipf scores
from \textit{wordfreq}~\citep{speer2022wordfreq} for
English; out-of-vocabulary tokens are assigned
to the \textsc{low} bucket.

\textbf{Decoder and evaluation.}
Unless noted otherwise, all reconstruction uses Gemini 2.0 Flash (zero-shot); BERTScore is computed with \texttt{distilroberta-base}~\citep{sanh2019distilbert} embeddings on English text and \texttt{bert-base-chinese}~\citep{devlin2019bert} on Chinese text. Prompts are in Appendix~\ref{app:prompts} and compute in Appendix~\ref{app:compute}.

\textbf{Compression ratio.}
Of the three storage tiers introduced in Section~\ref{sec:problem}, the main text reports tier 1, the skeleton itself: a fraction $\rkeep$ of the original characters plus ${<}0.1\%$ of encoding metadata (bucket weights or step parameters), excluding the shared decoder and prompt. Appendix~\ref{app:lossless_baselines} tabulates tier 2, the skeleton after lossless coding, which is the storage figure of merit and the only tier comparable with a lossless codec, and tier 3, the decoder's system cost (Table~\ref{tab:tiers}).

\subsection{Main Results}
\label{sec:main_results}

\begin{table*}[t]
\centering
\small
\setlength{\tabcolsep}{4.5pt}
\caption{All methods on the same $100$ BBC chunks, decoded with Gemini~2.5 Flash
(BERTScore F1, mean $\pm$ std; differences below about $0.003$, the $95\%$ confidence half-width at $n{=}100$, are ties).
WordFreq = WordFreq-3 deletion reconstructed with the same prompt as the other deletion methods. Entropy = GPT-2
surprisal-based token dropping. Summarization = compress-to-length rewriting.
\textbf{Bold} = best per row. $^{\dagger}$LLMLingua-2~\citep{pan2024llmlingua}, the classifier-based,
rate-targeted successor to LLMLingua~\citep{jiang2023llmlingua}, run as a drop-in encoder. Realised character
ratios differ by method (Appendix~\ref{app:lp_derivation}); full LLMLingua-2 comparison in Appendix~\ref{app:llmlingua2}.}
\label{tab:semantic_baselines}
\begin{tabular}{lcccccc}
\toprule
$\rkeep$ & WordFreq & Entropy & Entropy-LP & Hybrid-$0.5$ & Summ. & LLMLingua-2$^{\dagger}$ \\
\midrule
0.9 & $0.9815 \pm 0.008$ & $0.9881 \pm 0.012$ & $0.9893 \pm 0.007$ & $\mathbf{0.9912} \pm 0.006$ & $0.9628 \pm 0.017$ & $0.9857 \pm 0.007$ \\
0.7 & $0.9618 \pm 0.012$ & $0.9777 \pm 0.010$ & $\mathbf{0.9782} \pm 0.011$ & $0.9775 \pm 0.010$ & $0.9505 \pm 0.015$ & $0.9709 \pm 0.008$ \\
0.5 & $0.9382 \pm 0.014$ & $0.9531 \pm 0.013$ & $\mathbf{0.9543} \pm 0.012$ & $0.9524 \pm 0.013$ & $0.9255 \pm 0.012$ & $0.9462 \pm 0.010$ \\
0.3 & $0.9059 \pm 0.012$ & $0.9093 \pm 0.014$ & $0.9115 \pm 0.012$ & $0.9097 \pm 0.014$ & $0.8969 \pm 0.012$ & $\mathbf{0.9121} \pm 0.011$ \\
0.1 & $0.8752 \pm 0.011$ & $0.8735 \pm 0.010$ & $0.8735 \pm 0.010$ & $0.8721 \pm 0.011$ & $0.8636 \pm 0.014$ & $\mathbf{0.8805} \pm 0.009$ \\
\bottomrule
\end{tabular}
\end{table*}

\textbf{Character-level deletion: uniform vs.\ stochastic.} Table~\ref{tab:char_deletion} compares the four character-level rules and shows a clean regime shift. At $\rkeep \ge 0.7$ all four remain recoverable and Step (Uniform) is the strongest, so a regular deletion pattern preserves distributed context better than random deletion. From $\rkeep = 0.6$ Gaussian overtakes it ($0.943$ vs.\ $0.913$) and stays the best rule. Below $\rkeep = 0.5$ Step's BERTScore flattens near $0.86$ although the character error rate of the same outputs climbs from $0.80$ to $1.01$, {\em i.e.}, by $\rkeep = 0.1$ the output needs as many edits as the original has characters: the decoder writes fluent, largely invented text of about the original length. Even here the best rule depends on rate, but all four ignore word boundaries and semantic structure, so they are a lower-bound family: structure-aware deletion is needed once the goal is faithful reconstruction at useful compression rates.

\textbf{Word-level deletion methods.}
We next evaluate word-level deletion strategies, where
entire words are removed based on linguistic features.
Table~\ref{tab:unified} compares word-level deletion methods after LLM recovery.
The ranking is clearly regime-dependent. WordLen is strongest at mild compression, LP takes over in the middle regime, and WordFreq becomes best again at the most aggressive settings. Relative to the character-level baseline, all three word-level methods beat Step through the main middle range, but only WordFreq and LP remain reliably stronger once retention becomes very small. The main takeaway is therefore not that one word-level heuristic wins universally, but that the stronger schemes trade places as the operating point shifts. LP is informative for understanding bucket allocation, yet that does not consistently translate into the strongest end-to-end scores, which makes it more useful analytically than as a universally best practical choice.

Table~\ref{tab:unified} also reports ROUGE-L and entity recall, under which that conclusion survives. Read
alongside BERTScore, they locate a change in what reconstruction \emph{is}. At mild retention the
three track one another: the decoder is restoring the source. As retention falls they separate: semantic similarity holds up while lexical overlap and entity recall fall away, and that separation is the
signature of an output that is no longer a restoration of the original but a fluent substitute for it.
BERTScore cannot mark the boundary, since genre-consistent prose clears $0.8$ whatever it asserts: the Step
baseline flattens near $0.86$ while its ROUGE-L falls to $0.09$. The lexical and factual
lenses place that boundary.

They also separate strategies that look alike in aggregate. At $\rkeep = 0.5$ and $0.4$ LP leads on both similarity
metrics yet has the lowest entity recall of the three, because its deletion budget is spent against an
objective with no term for named entities: optimising a proxy improves the proxy, not the thing it stands
for. A strategy chosen on similarity alone can therefore be the wrong one wherever the reconstruction will be
read as a record. 

\begin{table}[b]
\centering
\small
\setlength{\tabcolsep}{4pt}
\caption{Encoder latency per 512-char chunk. Hybrid-$\alpha$ matches Entropy (GPT-2 inference dominates).}
\label{tab:latency}
\begin{tabular}{lcc}
\toprule
\textbf{Method} & \textbf{GPU (ms)} & \textbf{CPU (ms)} \\
\midrule
Step / WordFreq / LP & ${\sim}1$ & ${\sim}1$ \\
Entropy / Hybrid-$\alpha$ & ${\sim}15$ & ${\sim}130$ \\
\bottomrule
\end{tabular}
\end{table}

\begin{table*}[t]
\centering
\small
\setlength{\tabcolsep}{5pt}
\caption{Named-entity metrics computed on the \emph{reconstructions}. \emph{Entity recall} = original
entities present in the reconstruction (higher is better). \emph{Unsupported-entity rate} = reconstruction
entities absent from the original (lower is better). Word-level rows use the $200$ BBC test articles
($8{,}021$ entity mentions). Semantic rows use the $100$ $512$-character chunks, the same sample as
Table~\ref{tab:ner_preservation}, decoded with Gemini~2.5 Flash. \textbf{Bold} = best per column within a block. Recovery rates are
discussed in the text.}
\label{tab:ner_recon}
\begin{tabular}{l ccccc ccccc}
\toprule
& \multicolumn{5}{c}{\textbf{Entity recall}} & \multicolumn{5}{c}{\textbf{Unsupported-entity rate}} \\
\cmidrule(lr){2-6}\cmidrule(lr){7-11}
$\rkeep$ & 0.9 & 0.7 & 0.5 & 0.3 & 0.1 & 0.9 & 0.7 & 0.5 & 0.3 & 0.1 \\
\midrule
Step                   & .947 & .728 & .088 & .026 & .014 & .048 & .229 & .787 & .883 & .930 \\
WordLen                & \textbf{.978} & \textbf{.947} & \textbf{.760} & .264 & .023 & \textbf{.018} & \textbf{.041} & .171 & .551 & .878 \\
WordFreq-3             & .937 & .854 & .722 & \textbf{.544} & \textbf{.268} & .040 & .076 & \textbf{.133} & \textbf{.158} & \textbf{.237} \\
WordFreq-6             & .936 & .814 & .651 & .404 & .172 & .044 & .109 & .155 & .203 & .249 \\
LP                     & .954 & .883 & .653 & .398 & .191 & .026 & .066 & .145 & .176 & .264 \\
\midrule
Entropy                & .947 & .897 & .758 & .476 & \textbf{.177} & .052 & .079 & .163 & .300 & .538 \\
Entropy-LP             & .956 & .891 & .767 & .525 & \textbf{.177} & .056 & .078 & \textbf{.133} & \textbf{.278} & \textbf{.533} \\
Ent.-FreqBkt           & .974 & .665 & .310 & .189 & .072 & .051 & .198 & .389 & .430 & .646 \\
Hybrid-$\alpha{=}0.3$  & .979 & .911 & \textbf{.788} & .536 & .173 & .042 & .072 & .155 & .293 & .577 \\
Hybrid-$\alpha{=}0.5$  & \textbf{.988} & .938 & .786 & \textbf{.538} & .171 & .033 & \textbf{.051} & .153 & .284 & .583 \\
Hybrid-$\alpha{=}0.7$  & .986 & \textbf{.941} & .781 & .488 & .127 & \textbf{.029} & .060 & .172 & .326 & .678 \\
\bottomrule
\end{tabular}
\end{table*}

\begin{table}[t]
\centering
\small
\setlength{\tabcolsep}{4pt}
\caption{Fine-tuned Llama-3.2-3B-Instruct vs.\ Gemini 2.0 Flash (BERTScore F1). At each rate all three
decoders reconstruct the deletion strategy whose fine-tuned decoder performs best (LP at $\rkeep = 0.9$ and
$0.5$, WordLen at $0.7$, WordFreq-3 at $0.3$ and $0.1$); the Gemini values are the matching cells of
Table~\ref{tab:unified}. ``ZS'' = zero-shot Llama, ``SFT'' = QLoRA fine-tuned Llama.}
\label{tab:finetune}
\begin{tabular}{lcc}
\toprule
$\rkeep$ & Llama & Gemini \\
 & ZS / SFT & Flash \\
\midrule
0.9 & 0.9857 / \textbf{0.9967} & 0.9885 \\
0.7 & 0.9465 / 0.9608 & \textbf{0.9865} \\
0.5 & 0.9032 / \textbf{0.9598} & 0.9484 \\
0.3 & 0.8711 / \textbf{0.9206} & 0.8948 \\
0.1 & 0.8436 / \textbf{0.8827} & 0.8527 \\
\bottomrule
\end{tabular}
\end{table}

\textbf{Semantic Baselines.}\label{sec:new_baselines}
Table~\ref{tab:semantic_baselines} shows the main semantic-baseline comparison. Entropy-based deletion leads WordFreq by $0.007$--$0.016$ from $\rkeep = 0.9$ to $0.5$, the gap narrows to $0.003$ at $0.3$, and at $0.1$ WordFreq is level with the entropy family ($0.875$ against $0.874$), while LLM summarization is the weakest method at every rate: compress-to-length rewriting loses lexical detail that the skeleton-plus-reconstruction pipeline preserves. Two distinctions qualify entropy's advantage. First, entropy needs GPT-2 inference at the encoder, and Hybrid-$\alpha$ costs the same because that inference dominates, whereas WordFreq needs only a dictionary lookup (roughly $15\times$ faster on GPU and $130\times$ on CPU, Table~\ref{tab:latency}). The gap it buys is at most $0.016$ BERTScore, so WordFreq is the practical choice when encoder resources are constrained. That latency buys fewer bytes stored and transmitted (Appendix~\ref{app:lossless_baselines}), not faster downstream inference, since a consumer that reconstructs receives a text of roughly the original length $L$ again. Second, summarization rewrites the text entirely and can lose named entities and exact phrasing, whereas deletion keeps original tokens as anchors. These observations motivate hybrids that keep the skeleton paradigm but use stronger importance signals than frequency alone.

\textbf{Hybrid deletion methods.}
\label{sec:hybrid}
Table~\ref{tab:semantic_baselines} gives a compact view of the semantic methods and representative hybrids. Three patterns matter. First, Entropy-LP matches or beats pure entropy at every rate, indicating that LP-style budget allocation remains helpful when the buckets are defined by surprisal rather than Zipf frequency. Second, Entropy-in-FreqBuckets is weaker than pure entropy, suggesting that frequency-derived bucket boundaries are not a good structural match for entropy-based token selection. Third, Hybrid-$\alpha$ leads only at mild-to-moderate compression, so combining frequency and surprisal pays when encoder-side GPT-2 inference is affordable, while WordFreq stays hard to beat at the sparsest rates: extra encoder-side computation buys its clearest gains where the skeleton is sparse enough for token importance to matter but not so sparse that every method is dominated by decoder uncertainty. Table~\ref{tab:semantic_baselines} reports BERTScore only. The factual and lexical views of the same methods (Tables~\ref{tab:ner_recon} and~\ref{tab:rougel_semantic}) qualify this ranking rather than overturn it: Entropy-LP leads on BERTScore and ROUGE-L at $\rkeep = 0.7$, where entity recall still favours Hybrid-$\alpha{=}0.5$. The hybrid family is therefore not uniformly superior. Its value is a sweet spot different from the low-resource baseline's.

\textbf{LLMLingua-2 as an encoder.}
A learned prompt compressor does not dominate the handcrafted heuristics on the reconstruction
objective: in Table~\ref{tab:semantic_baselines} it trails Entropy-LP from
$\rkeep = 0.9$ to $0.5$, ties it at $0.3$ and leads the table at $0.1$, by $0.005$. Appendix~\ref{app:llmlingua2} gives the full comparison, including what changes when
the compressed text is consumed without reconstruction.

\textbf{Fine-Tuning Results.}
\label{sec:finetune_results}
Table~\ref{tab:finetune} isolates the decoder under the same skeletons. QLoRA fine-tuning of Llama-3.2-3B-Instruct improves over zero-shot Llama at every rate and beats Gemini at four of the five. Only at $\rkeep = 0.7$ does Gemini stay ahead. At high retention the gain is fluency repair that brings a 3B decoder near Gemini quality. At $\rkeep = 0.3$ and $0.1$ it is structural, since recovery from a sparse skeleton depends on learning the deletion pattern, and these are exactly the settings where a lightweight local decoder would otherwise be least reliable. Encoder design and decoder adaptation are complementary levers on the paired comparison of Appendix~\ref{app:levers}: with the skeleton fixed, QLoRA gains $0.011$--$0.056$ BERTScore. With the decoder fixed, the choice of word-level rule spans $0.005$--$0.069$, and the encoder lever is the larger one at $\rkeep = 0.7$ and $0.1$ (Appendix~\ref{app:levers}). The comparison is not an architecture ranking between Llama and Gemini, and it is kept apart from Tables~\ref{tab:unified} and~\ref{tab:semantic_baselines}, which fix the decoder, because it changes the decoder itself.

\section{Analysis and Discussion}
\label{sec:analysis}

\textbf{Failure modes and factual drift.}
Aggregate similarity tells us how far a reconstruction lands from its source, not how it goes wrong. Two decoder failure modes recur once retention is low: output repetition is the most common Llama failure at $\rkeep \le 0.2$, and Gemini occasionally exceeds the requested length budget at very low retention. Both leave the output fluent, so neither is penalised much by BERTScore, the reason we quantify factual fidelity separately. Appendix~\ref{app:examples} shows reconstructions at several rates for qualitative reference. We therefore track named entities with spaCy~\citep{honnibal2020spacy} named-entity recognition (NER), first in the skeleton alone. The per-rate survival figures are in Appendix~\ref{app:detailed_metrics}, and three patterns stand out. First, uniform character deletion is especially damaging for factual anchors: even when BERTScore remains superficially acceptable, the skeleton quickly loses named entities. Second, WordFreq is the strongest purely frequency-based method across most of the range, while the best hybrids match it at mild compression and lead it at $\rkeep = 0.3$. This is an important complement to the BERTScore results, because it shows that some of the hybrid advantage is not merely stylistic fluency but better retention of factual anchor tokens. Third, at the most aggressive settings every skeleton loses most of its entities, the semantic ones slightly fewer, but that advantage vanishes after reconstruction: WordFreq recalls more entities than any semantic method at $\rkeep = 0.1$ (Table~\ref{tab:ner_recon}) and twice as many, at a third of the unsupported-entity rate, at $\rkeep = 0.05$ (Appendix~\ref{app:r005}). WordLen is a useful edge case: it preserves entities unusually well at very high retention because its early edits mainly affect spaces and internal vowels, but that advantage collapses once compression becomes more severe: at $\rkeep = 0.9$ its only edit is whitespace, so every entity survives intact, but from $\rkeep = 0.7$ vowel deletion and stem truncation damage the entity strings themselves, and only the decoder can restore them.

\textbf{Entities after reconstruction.}
Skeleton survival bounds what the decoder is given but says nothing about what it returns. We therefore score the reconstructions directly with the same NER
pipeline, using three metrics (Table~\ref{tab:ner_recon}):
\emph{entity recall}, the share of original entities present in the reconstruction; \emph{recovery rate},
the share of skeleton-deleted entities the decoder restores; and \emph{unsupported-entity rate}, the share
of entities in the reconstruction with no match in the original. We reserve the term \emph{hallucinated
claim} for the human and LLM-rater judgments of Appendix~\ref{app:fact_eval}: an unsupported entity is
usually a distorted version of a real one rather than an invention, and only a reader comparing the two
texts can tell those apart.

The decoder is a genuine second chance, and how much a strategy needs it separates two families.
On the $200$-article sample of Table~\ref{tab:ner_recon}, WordLen at $\rkeep = 0.5$ keeps only $13.5\%$ of entities in its skeleton yet reaches $76.0\%$ recall
after reconstruction, restoring $73.1\%$ of what it deleted: its edits hit spaces and internal vowels, leaving damaged but inferable tokens. WordFreq is the mirror image: at the same rate it keeps $61.6\%$ in
the skeleton and recalls $72.2\%$, recovering only $32.3\%$ of what it deleted. The two reach comparable
factual fidelity by opposite routes: one carries its anchors in the skeleton, the other rebuilds them. Only the first route still works when the decoder is weak, which is why WordFreq remains the safer
default at aggressive rates. Step has no such route: below $\rkeep = 0.7$ its recall collapses to $1.4$--$8.8\%$.

Reconstruction is not free, though, and the cost is visible in the third metric. Unsupported entities stay
at or below $5\%$ at $\rkeep = 0.9$ for every word-level strategy, but rise to $13$--$17\%$ at $\rkeep = 0.5$ and to
$24$--$88\%$ at $\rkeep = 0.1$, with WordLen the extreme case: the same mechanism that lets it rebuild
entities from damaged tokens also lets it invent them once the damage is too severe. This is the
quantitative form of the BERTScore leniency noted above: a reconstruction can read fluently, score
above $0.8$, and still populate itself with entities that were never in the source.

The lowest-retention regime is therefore best interpreted as a stress test of decoder robustness rather than a generally practical operating point. Table~\ref{tab:regimes} maps retention rates to factual-error bands and the uses they support, Appendix~\ref{app:r005} extends the grid to $\rkeep = 0.05$, and Appendix~\ref{app:choosing} turns the results into a default rule for choosing a strategy on a new domain. A blind human study of fact preservation (Appendix~\ref{app:human_eval}), scaled to all strategies and rates by two LLM raters calibrated against those annotators (Appendix~\ref{app:fact_eval}), confirms this reading: fact preservation stays at or above $0.9$ at $\rkeep=0.9$ for every word-level strategy, WordFreq and LP are the only ones retaining facts at $\rkeep\le0.2$, and invented claims remain rare ($\le4\%$ of items for WordLen, WordFreq and LP) at $\rkeep\ge0.7$.

\textbf{Cross-Dataset Perspective.}
The BBC News results above establish the main picture in a controlled factual domain, but the broader evaluation reveals which conclusions appear robust and which depend on dataset or language. Across the English datasets, including Wikipedia (Appendix~\ref{app:wikipedia_eval}) and Reddit (Appendix~\ref{app:reddit_eval}), several findings recur. Uniform step deletion is consistently weak once compression becomes substantial, confirming that indiscriminate dropping removes too much local structure for reliable reconstruction. Frequency-based deletion remains a strong low-cost baseline across domains, even when it is not always the best method at a given operating point. At the same time, no single deletion rule dominates everywhere: semantic and hybrid methods help in specific regimes rather than uniformly. The broader evaluations show where the picture changes. Entity-rich news text rewards methods that preserve factual anchors; encyclopedic text more often benefits from semantic bucket allocation; conversational text makes contextual surprisal more useful; and the Chinese datasets show that the strongest method can shift with register, segmentation, and information density. These domain shifts track intuitively with differences in discourse structure, lexical repetition, and how much information is concentrated in a small number of rare tokens. The appendix therefore serves as a structured extension of the main BBC analysis: a map of which conclusions transfer and which remain domain-dependent.

This cross-dataset view also clarifies where lossy semantic compression is most relevant in practice. Typical lossless text codecs already achieve moderate compression, so the practical motivation for a lossy semantic method is strongest once the target retention rate falls meaningfully below the range where exact recovery is still realistic. In that lower-retention regime, the objective changes: the goal is no longer byte-exact reconstruction, but preserving enough meaning, factual anchors, and textual structure for useful downstream recovery. This is also the regime in which the choice of deletion strategy matters most. When compression is mild, many methods remain competitive and pure lossless coding is often still a viable alternative; when the retained text must be much smaller, differences between frequency-based, entropy-based, and hybrid deletion become more consequential. At the same time, the two approaches are complementary rather than exclusive: a retained semantic skeleton can still be compressed further with a conventional lossless codec. The broader evaluations reported in the appendix are intended to sharpen that practical picture by showing how these trade-offs evolve across additional English and Chinese domains.

\section{Conclusions}
\label{sec:conclusion}

We studied lossy semantic text compression via token deletion and LLM reconstruction, benchmarking frequency-based, entropy-based, and hybrid deletion rules and scoring the results for semantic similarity, lexical overlap, and factual fidelity. Three conclusions emerge. First, frequency-guided deletion remains a strong and efficient baseline: a static lookup is competitive with neural encoders at mild compression and is the safest of our rules at the lowest retention rate, because the entities it keeps ride in the skeleton rather than being regenerated. Second, semantic and hybrid methods pay in the middle regime, where the skeleton is sparse enough for token importance to matter, and once it becomes sparser LP-style allocation over surprisal buckets holds up better than direct rank interpolation. Third, BERTScore alone overstates what survives: below about half retention, reconstructions stay fluent while entity recall falls and unsupported entities rise, so the low-retention regime is a stress test rather than an operating point, and the retention bands of Table~\ref{tab:regimes} are the practical rule. Decoder adaptation is a separate axis: QLoRA fine-tuning gives a compact local decoder competitive with a commercial API. Across English and Chinese domains the framework transfers, but the strongest deletion rule still depends on domain, language, and operating regime.

\section{Limitations}
\label{sec:limitations}

Primary experiments are on English news text; the Wikipedia, Reddit, and Chinese results
are encouraging but literary, technical, code-heavy, and privacy-sensitive domains remain open.
The appendix covers all method families, including entropy, hybrid, and SFT results,
across the evaluated English and Chinese datasets, but these are still public-domain benchmarks.
This framework is therefore most plausible for familiar, repeated, or conventionally structured
factual text, where the decoder's pretrained world knowledge and stylistic priors can help fill
in gaps from a sparse skeleton. Public-domain overlap may make reconstruction easier in these
settings, so strong performance should not be over-interpreted as guaranteed robustness on private,
specialized, or highly novel corpora.

Metric comparability is also limited: English BERTScore uses distilroberta-base, and Chinese
BERTScore uses bert-base-chinese, so scores are not directly
comparable across languages and datasets. Human fact-level judgments cover only two strategies at three
retention rates (Appendix~\ref{app:human_eval}). The wider grid relies on LLM raters and automatic proxies
(Appendices~\ref{app:fact_eval} and~\ref{app:detailed_metrics}), and even BERTScore plus lexical metrics
can be too forgiving at very low retention, since fluent but factually incorrect reconstructions still
receive nontrivial similarity scores.

This framework is also not intended as a default solution for high-stakes text domains
where exact fidelity is essential. In legal, medical, financial, compliance, or audit-sensitive
settings, even small reconstruction errors in names, numbers, dates, negation cues, or contractual
wording may be unacceptable. Our method is lossy by design: although it often preserves much of the
original meaning, it does not provide guarantees of exact recovery and can fail more severely at low
retention rates. For such applications, lossless storage or transmission remains the safer default.

Even everyday text carries a related risk, {\em e.g.}, in agent memory, one of our motivating settings. When an interaction history or a
user-written note is compressed and cached, only the reconstruction is read back later. The original is no longer present, so no reader can make the side-by-side comparison our annotators needed to detect
distortion (Appendix~\ref{app:human_eval}). Reconstruction errors are fluent, plausible substitutions
rather than visible damage: a name, number, or negation is replaced by a grammatical alternative of the
same kind, and the result still reads as a record. That is precisely what makes them risky once the
reconstructed text is treated as a faithful account of what a user said or agreed to. The decoder can also
fill a gap with pretrained knowledge about a real person or event that the source never contained. Where
reconstructions are stored as memory, we recommend the mild regime only (Table~\ref{tab:regimes}; entity
recall $0.89$--$0.95$), keeping user-provided names, quantities, and commitments losslessly alongside the
skeleton, and labelling reconstructed text as reconstructed rather than as the original.

LLM decompression adds $1$--$3$\,s of latency per chunk whenever the full text is recovered, which rules out
real-time access to reconstructions. Consuming the skeleton directly avoids the cost but is viable only for
word-level strategies (Appendix~\ref{app:downstream}). The
lowest-retention regime should be read mainly as a stress test of decoder failure rather than as a
generally practical operating point. The LP assumes bucket independence and therefore cannot model
token interactions directly. Fine-tuning remains decoder- and training-distribution-dependent:
it is strongest on BBC News, Reddit, and Zhihu, but less competitive on Chinese news and Chinese
Wikipedia. While the downstream evaluation (Appendix~\ref{app:downstream}) covers two tasks on
one dataset, our focus is reconstruction fidelity for reusable compressed text artifacts rather than
task-specific prompt compression. Finally, the zero-shot decoder is a proprietary API model, and its version was retired during review. The affected additions were run on the successor, which reproduces the published aggregates within $0.016$~F1 (Table~\ref{tab:decoder_check}), and all per-sample outputs are released (Appendix~\ref{app:reproducibility}) so that the evaluation can be reproduced without the original API.

\section*{Acknowledgments}

We thank Nageen Saira (CUNY Queens College) for serving as an independent annotator in the
human fact-preservation evaluation, and the ARR reviewers and area chair for feedback that
shaped the camera-ready revision.

\paragraph{AI assistance disclosure.}
LLMs were used in this project in three roles. First, proprietary and open-source LLMs were used as
\emph{experimental decoders} in the benchmark itself: compressed text skeletons were reconstructed
by Gemini, Llama-3.2-3B-Instruct, and Qwen2.5-3B-Instruct, and then evaluated quantitatively.
In the scaled fact-level evaluation (Appendix~\ref{app:fact_eval}), Gemini also drafted the facts for
texts outside the human study, and Claude Opus~5 and GPT-5.6 served as raters.
These model outputs are part of the reported experimental results and therefore constitute disclosed
LLM use in evaluation. Second, AI coding assistants were used to help write, debug, and re-run
analysis scripts and to check that reported numbers match the stored result files. The authors
reviewed the code and verified every reported number. Third, LLMs were used during manuscript
preparation to help draft and revise prose passages; any such text was subsequently checked, edited, and
approved by the authors before inclusion. LLMs were \emph{not} used to originate the core research
idea, generate the underlying datasets, or produce figures without author verification. All
methodological decisions, experiment design, data selection, result validation, interpretation, and
final writing decisions were made by the authors. The authors take full responsibility for the
content of the paper, including the accuracy of reported results, the correctness of claims, and
any errors that may remain. No LLM is listed as an author or regarded as having made an
authorship-level intellectual contribution.

\bibliography{references}

\appendix

\section{Detailed Metrics}
\label{app:detailed_metrics}

BERTScore carries most of the paper's conclusions, and it can reward a fluent, genre-consistent
reconstruction whose content has largely gone. This section reports the metrics we use to check it on the
main English benchmark, BBC News: per-rate BERTScore statistics with confidence intervals for WordFreq, character error rate (CER) for all four word-level strategies, ROUGE-L for the same four strategies, and chrF++~\citep{popovic2017chrf} and METEOR~\citep{banerjee2005meteor} with a rank-stability analysis. CER is
the normalized Levenshtein distance between the reconstruction and the original at character level (lower is better). ROUGE-L is the longest-common-subsequence F-measure (higher is better). All of these are
lexical or surface metrics, so unlike BERTScore they do not reward fluent paraphrase or genre-consistent
filler. We did not add COMET, which is trained on human judgments of machine translation and is not
validated for same-language reconstruction, or perplexity, which measures fluency rather than fidelity and
would reward exactly the failure mode we are trying to detect.  The section closes with named-entity survival in the skeleton before reconstruction.

\begin{table*}[tbp]
\centering
\small
\caption{Per-rate BERTScore F1 for the default WordFreq (3-class) strategy on BBC News ($n{=}200$ chunks):
mean, standard deviation, and 95\% confidence interval ($\bar{x} \pm 1.96\, s/\sqrt{n}$). CER for this and the
other word-level strategies is in Table~\ref{tab:cer_detail}.}
\label{tab:wordfreq_detail}
\begin{tabular}{lccc}
\toprule
$\rkeep$ & Mean F1 & Std & 95\% CI \\
\midrule
0.90 & 0.9839 & 0.0052 & $[0.983, 0.985]$ \\
0.80 & 0.9754 & 0.0071 & $[0.974, 0.976]$ \\
0.70 & 0.9645 & 0.0092 & $[0.963, 0.966]$ \\
0.60 & 0.9479 & 0.0121 & $[0.946, 0.950]$ \\
0.50 & 0.9314 & 0.0136 & $[0.930, 0.933]$ \\
0.40 & 0.9137 & 0.0143 & $[0.912, 0.916]$ \\
0.30 & 0.8948 & 0.0135 & $[0.893, 0.897]$ \\
0.20 & 0.8737 & 0.0141 & $[0.872, 0.876]$ \\
0.10 & 0.8527 & 0.0136 & $[0.851, 0.855]$ \\
\bottomrule
\end{tabular}
\end{table*}

Table~\ref{tab:wordfreq_detail} gives the per-rate dispersion behind the WordFreq column of
Table~\ref{tab:unified}. Two things are visible. First, the per-chunk standard deviation grows as
compression tightens, from $0.005$ at $\rkeep = 0.9$ to about $0.014$ from $\rkeep = 0.4$ downward:
at mild compression nearly every chunk is restored equally well, whereas at aggressive rates the outcome
depends on the chunk, since some skeletons keep the anchors the decoder needs and others do not. The
dispersion plateaus rather than growing further, which suggests that below $\rkeep = 0.4$ the
decoder's success is governed by chunk content rather than by the exact retention rate. Second, because
$n = 200$, the $95\%$ confidence half-width stays at or below $0.003$ F1 at every rate despite that
dispersion. The differences of $0.01$--$0.03$ between word-level strategies in
Table~\ref{tab:unified} are therefore well outside sampling noise, while gaps below about $0.005$,
such as the $0.0024$ between LP and WordFreq at $\rkeep = 0.3$, are ties that the lexical metrics
must break.

\begin{table*}[tbp]
\centering
\small
\caption{Character error rate (CER, lower is better; mean over test chunks) for the four BBC News
word-level deletion methods. CER can exceed $1$ when the reconstruction is longer than the original.
\textbf{Bold} = lowest CER per row.}
\label{tab:cer_detail}
\begin{tabular}{lcccc}
\toprule
$\rkeep$ & Step & WordLen & WordFreq & LP \\
\midrule
0.90 & \textbf{0.059} & 0.067 & 0.104 & 0.063 \\
0.80 & 0.065 & \textbf{0.062} & 0.160 & 0.084 \\
0.70 & 0.195 & \textbf{0.069} & 0.227 & 0.155 \\
0.60 & 0.631 & \textbf{0.092} & 0.307 & 0.235 \\
0.50 & 0.607 & \textbf{0.246} & 0.400 & 0.335 \\
0.40 & 1.210 & \textbf{0.359} & 0.501 & 0.460 \\
0.30 & 1.052 & \textbf{0.606} & 0.615 & 0.610 \\
0.20 & 1.022 & \textbf{0.727} & 0.733 & 0.758 \\
0.10 & 0.920 & \textbf{0.833} & 0.851 & 0.894 \\
\bottomrule
\end{tabular}
\end{table*}

Table~\ref{tab:cer_detail} turns to lexical fidelity, where the picture differs from BERTScore. For
WordFreq, CER grows roughly linearly as retention falls from $\rkeep = 0.9$ to $0.5$ (about $0.07$ per
step) and then accelerates below $\rkeep = 0.5$ (about $0.11$ per step): once most of the text is gone,
the decoder fills in plausible but different wording, which BERTScore forgives and CER does not.
Across strategies, WordLen has the lowest CER through
$\rkeep = 0.4$ because its early edits (whitespace, vowels, short words) leave long verbatim runs that
the decoder restores exactly, and Step's CER rises to $0.6$ by $\rkeep = 0.5$ while the BERTScore of the same outputs still
reads $0.83$. At $\rkeep \le 0.3$ all four strategies exceed
$0.6$ CER, {\em i.e.}, most characters of the reconstruction differ from the original even when BERTScore
remains above $0.80$. This is the regime the fact-level evaluation in Appendix~\ref{app:fact_eval}
characterizes directly.

\begin{table*}[tbp]
\centering
\small
\caption{ROUGE-L F-measure (mean $\pm$ std) for the four BBC News word-level deletion methods ($n{=}200$).}
\label{tab:rougel}
\begin{tabular}{lcccc}
\toprule
$\rkeep$ & Step & WordLen & WordFreq & LP \\
\midrule
0.90 & $0.952 \pm 0.056$ & $0.950 \pm 0.062$ & $0.923 \pm 0.045$ & $\mathbf{0.954 \pm 0.045}$ \\
0.80 & $0.941 \pm 0.051$ & $\mathbf{0.953 \pm 0.052}$ & $0.878 \pm 0.045$ & $0.940 \pm 0.042$ \\
0.70 & $0.772 \pm 0.100$ & $\mathbf{0.942 \pm 0.081}$ & $0.824 \pm 0.047$ & $0.889 \pm 0.050$ \\
0.60 & $0.414 \pm 0.213$ & $\mathbf{0.918 \pm 0.045}$ & $0.752 \pm 0.051$ & $0.826 \pm 0.057$ \\
0.50 & $0.149 \pm 0.058$ & $0.714 \pm 0.092$ & $0.664 \pm 0.050$ & $\mathbf{0.748 \pm 0.054}$ \\
0.40 & $0.084 \pm 0.048$ & $0.588 \pm 0.090$ & $0.568 \pm 0.052$ & $\mathbf{0.628 \pm 0.058}$ \\
0.30 & $0.074 \pm 0.042$ & $0.210 \pm 0.071$ & $0.449 \pm 0.050$ & $\mathbf{0.478 \pm 0.061}$ \\
0.20 & $0.075 \pm 0.036$ & $0.111 \pm 0.033$ & $\mathbf{0.323 \pm 0.050}$ & $0.312 \pm 0.052$ \\
0.10 & $0.054 \pm 0.024$ & $0.073 \pm 0.027$ & $\mathbf{0.191 \pm 0.046}$ & $0.142 \pm 0.035$ \\
\bottomrule
\end{tabular}
\end{table*}

\begin{table*}[tbp]
\centering
\small
\caption{ROUGE-L F-measure for the semantic and hybrid strategies, on the $100$ BBC chunks used in
Table~\ref{tab:semantic_baselines}, decoded with Gemini~2.5 Flash (means; $n{=}100$). \textbf{Bold} = best per column. Entity recall and
the unsupported-entity rate for the same methods are in Table~\ref{tab:ner_recon}.}
\label{tab:rougel_semantic}
\begin{tabular}{lccccc}
\toprule
$\rkeep$ & 0.9 & 0.7 & 0.5 & 0.3 & 0.1 \\
\midrule
Entropy                 & 0.942 & 0.883 & 0.703 & 0.392 & 0.154 \\
Entropy-LP              & 0.951 & \textbf{0.889} & \textbf{0.714} & \textbf{0.404} & \textbf{0.163} \\
Entropy-in-FreqBuckets  & \textbf{0.956} & 0.829 & 0.603 & 0.370 & 0.159 \\
Hybrid-$\alpha{=}0.3$   & 0.950 & 0.871 & 0.688 & 0.392 & 0.146 \\
Hybrid-$\alpha{=}0.5$   & \textbf{0.956} & 0.860 & 0.659 & 0.367 & 0.144 \\
Hybrid-$\alpha{=}0.7$   & 0.951 & 0.859 & 0.612 & 0.306 & 0.132 \\
\bottomrule
\end{tabular}
\end{table*}

Table~\ref{tab:rougel_semantic} extends ROUGE-L to the semantic and hybrid strategies on the $100$-chunk
sample. The ROUGE-L results in Table~\ref{tab:rougel} reinforce the main BERTScore story but sharpen the regime split. WordLen is strongest at $\rkeep = 0.8$--$0.6$, LP is best at $\rkeep = 0.9$ and $0.5$--$0.3$, and WordFreq is best again at the two most aggressive settings ($\rkeep = 0.2, 0.1$). Step remains much weaker once the deletion rate becomes substantial, collapsing especially at $\rkeep \le 0.5$. As expected, ROUGE-L is more sensitive than BERTScore to exact lexical overlap, so all methods degrade faster once the decoder must paraphrase or invent larger spans. These rows are reproduced directly from the stored BBC result files.

Table~\ref{tab:chrf_meteor} gives chrF++ and METEOR per strategy and rate. Their value is that they
refuse to saturate: WordLen at $\rkeep = 0.3$ retains a BERTScore of $0.85$ while its chrF++ is $0.286$ and
its METEOR $0.165$. The semantic metric reports a text of the same kind. The surface metrics report that
little of the wording survived. This is the same floor effect quantified against human judgment in
Appendix~\ref{app:human_eval}, and it is why Table~\ref{tab:unified} reports a lexical metric beside
BERTScore rather than relegating it here.

Table~\ref{tab:rank_stability} asks the complementary question, whether the choice of metric changes which
strategy wins. Spearman correlation between each metric's strategy ranking and BERTScore's is
$0.80$--$1.00$ for ROUGE-L across all nine rates. chrF++ and METEOR track BERTScore exactly from
$\rkeep = 0.8$ to $0.6$ ($\rho = 1.00$) but fall to $0.40$ at $\rkeep = 0.1$, their weakest point. The dips do not reflect disagreement about conclusions. All four metrics
select the same best strategy at five of the nine rates, and the four exceptions ($\rkeep = 0.9$, $0.5$,
$0.4$, $0.3$) are precisely the rates at which the top two strategies are within $0.004$ BERTScore of each
other, from $0.0001$ at $\rkeep = 0.9$ to $0.0035$ at $0.4$. At margins that small the ordering sits inside
the benchmark's own sampling noise: the WordFreq confidence intervals above have half-width
${\le}0.003$~F1, wider than three of these four gaps. Where the margin is larger the metrics agree: at $\rkeep = 0.2$ and $0.1$, which include the weakest point of chrF++ and METEOR, all four metrics still select WordFreq.

Table~\ref{tab:ner_preservation} reports named-entity survival in the compressed skeleton before reconstruction, the skeleton-side counterpart of Table~\ref{tab:ner_recon}: the share of the original's entity mentions that survive deletion, before the decoder restores anything.

\begin{table*}[tbp]
\centering
\caption{Named-entity survival in the compressed skeleton, \emph{before} reconstruction, measured with spaCy NER over $100$ test chunks of $512$ characters ($773$ named-entity mentions). Higher is better. Table~\ref{tab:ner_recon} reports what survives \emph{after} reconstruction.}
\label{tab:ner_preservation}
\small
\setlength{\tabcolsep}{6pt}
\begin{tabular}{lccccc}
\toprule
\textbf{Method} & $\rkeep{=}0.9$ & $\rkeep{=}0.7$ & $\rkeep{=}0.5$ & $\rkeep{=}0.3$ & $\rkeep{=}0.1$ \\
\midrule
Step & $0.396$ & $0.061$ & $0.016$ & $0.004$ & $0.005$ \\
WordLen & $\mathbf{1.000}$ & $0.312$ & $0.176$ & $0.019$ & $0.003$ \\
WordFreq & $0.930$ & $\mathbf{0.877}$ & $\mathbf{0.662}$ & $0.301$ & $0.038$ \\
LP & $0.838$ & $0.468$ & $0.204$ & $0.179$ & $0.101$ \\
Entropy & $0.825$ & $0.686$ & $0.558$ & $0.353$ & $0.128$ \\
Ent.-LP & $0.841$ & $0.691$ & $0.558$ & $0.362$ & $\mathbf{0.140}$ \\
Ent.-FreqBkt & $0.895$ & $0.485$ & $0.202$ & $0.148$ & $0.058$ \\
Hybrid-$0.3$ & $0.911$ & $0.726$ & $0.593$ & $0.396$ & $0.126$ \\
Hybrid-$0.5$ & $0.930$ & $0.824$ & $0.617$ & $\mathbf{0.409}$ & $0.126$ \\
Hybrid-$0.7$ & $0.933$ & $0.872$ & $0.651$ & $0.348$ & $0.092$ \\
\bottomrule
\end{tabular}
\end{table*}

\begin{table*}[tbp]
\centering
\small
\setlength{\tabcolsep}{4.5pt}
\caption{chrF++ and METEOR for the word-level strategies on BBC News, on the reconstructions scored in
Table~\ref{tab:unified}. \textbf{Bold} = best per column. }
\label{tab:chrf_meteor}
\begin{tabular}{lccccccccc}
\toprule
$\rkeep$ & 0.9 & 0.8 & 0.7 & 0.6 & 0.5 & 0.4 & 0.3 & 0.2 & 0.1 \\
\midrule
\multicolumn{10}{l}{\emph{chrF++}} \\
WordLen            & \textbf{.922} & \textbf{.924} & \textbf{.913} & \textbf{.886} & \textbf{.710} & \textbf{.580} & .286 & .181 & .094 \\
WordFreq-3         & .875 & .816 & .754 & .672 & .582 & .479 & .358 & \textbf{.233} & \textbf{.118} \\
WordFreq-6         & .882 & .823 & .749 & .659 & .543 & .408 & .277 & .171 & .079 \\
LP                 & .921 & .896 & .827 & .748 & .651 & .521 & \textbf{.360} & .203 & .078 \\
\addlinespace[3pt]
\multicolumn{10}{l}{\emph{METEOR}} \\
WordLen            & .908 & \textbf{.912} & \textbf{.901} & \textbf{.874} & \textbf{.649} & .465 & .165 & .091 & .049 \\
WordFreq-3         & .866 & .801 & .726 & .624 & .515 & .395 & .275 & \textbf{.161} & \textbf{.074} \\
WordFreq-6         & .875 & .808 & .719 & .604 & .465 & .319 & .196 & .111 & .048 \\
LP                 & \textbf{.910} & .890 & .832 & .754 & .649 & \textbf{.484} & \textbf{.295} & .141 & .041 \\
\bottomrule
\end{tabular}
\end{table*}

\begin{table}[htbp]
\centering
\small
\setlength{\tabcolsep}{4pt}
\caption{Spearman correlation between each metric's ranking of the four word-level configurations and
BERTScore's, per retention rate. A value of $1.00$ means the metric orders the strategies exactly as
BERTScore does.}
\label{tab:rank_stability}
\begin{tabular}{lccc}
\toprule
$\rkeep$ & ROUGE-L & chrF++ & METEOR \\
\midrule
0.9 & 1.00 & 0.80 & 1.00 \\
0.8 & 1.00 & 1.00 & 1.00 \\
0.7 & 1.00 & 1.00 & 1.00 \\
0.6 & 1.00 & 1.00 & 1.00 \\
0.5 & 1.00 & 0.80 & 0.80 \\
0.4 & 1.00 & 0.80 & 1.00 \\
0.3 & 0.80 & 0.60 & 0.80 \\
0.2 & 1.00 & 0.80 & 1.00 \\
0.1 & 0.80 & 0.40 & 0.40 \\
\bottomrule
\end{tabular}
\end{table}

\section{Human Fact-Level Evaluation}
\label{app:human_eval}

Entity matching and BERTScore are both automatic proxies. To check them against human judgment we ran a
blind study of $54$ reconstructions (\{WordFreq-6, LP\} $\times$ $\rkeep \in \{0.5, 0.3, 0.1\}$). Four annotators (three authors and one independent annotator not otherwise involved in the project) labelled each of the $3$--$5$ facts per item as preserved, distorted, or missing, and flagged whether the
reconstruction asserted anything invented, blind to strategy and rate throughout ($1{,}062$ judgments).
Inter-annotator agreement is $\kappa = 0.40$--$0.59$ three-way and $0.49$--$0.79$ on the binary
preserved-versus-not collapse, with most of the disagreement on the distorted-versus-missing boundary
rather than on whether a fact survived. Annotators were unpaid and saw only the two texts and the facts. All texts are publicly available BBC News articles, and no personal data about annotators or third parties
were collected.

Table~\ref{tab:human_eval} shows how little survives. Half the facts are intact at $\rkeep = 0.5$. By $\rkeep = 0.1$ almost none are ($3\%$ pooled), and what is lost is mostly simply absent rather than wrong.
The result that matters for the main text is the calibration. Fact preservation does correlate with
BERTScore across the $54$ items (Spearman $\rho = 0.776$, $p < 10^{-11}$), so the metric is not uninformative, but the two part company exactly where the paper's claims are weakest: at $\rkeep = 0.1$
these reconstructions still score near $0.84$ BERTScore while $3\%$ of their facts survive. That is the
human counterpart of the floor visible in Table~\ref{tab:unified}, and it is specific to BERTScore. ROUGE-L on the same items falls to $0.14$. The study also supplies a practical ordering the similarity metrics do
not: at $\rkeep = 0.5$, where both strategies are still usable, LP preserves more facts than WordFreq-6
($58\%$ against $47\%$).

Outright invention is rarer than the automatic metrics imply. Every item flagged by two or more annotators
was adjudicated jointly, and $4$ of the $54$ were confirmed as asserting something with no basis in the
original, three of them at $\rkeep = 0.1$. What the unsupported-entity rate of Table~\ref{tab:ner_recon}
mostly counts, then, is distortion of real entities rather than fabrication, which is why we keep the
automatic and human terms distinct.

\begin{table}[htbp]
\centering
\small
\setlength{\tabcolsep}{4pt}
\caption{Blind human fact-level evaluation ($54$ reconstructions, four annotators, $1{,}062$ judgments).
Share of facts judged preserved (P), distorted (D), or missing (M), averaged over annotators.}
\label{tab:human_eval}
\begin{tabular}{lcccccc}
\toprule
& \multicolumn{3}{c}{\textbf{WordFreq-6}} & \multicolumn{3}{c}{\textbf{LP}} \\
\cmidrule(lr){2-4}\cmidrule(lr){5-7}
$\rkeep$ & P & D & M & P & D & M \\
\midrule
0.5 & .47 & .35 & .18 & \textbf{.58} & .29 & .13 \\
0.3 & .16 & .31 & .54 & \textbf{.26} & .43 & .32 \\
0.1 & .01 & .12 & .88 & \textbf{.05} & .24 & .71 \\
\bottomrule
\end{tabular}
\end{table}

\section{Scaled Fact-Level Evaluation with LLM Raters}
\label{app:fact_eval}

BERTScore and entity survival do not directly measure whether the \emph{facts} of a text survive
reconstruction. We therefore scale the blind fact-level protocol of our human study to the full
strategy--rate grid using two LLM raters from different model families than the decoder.

\paragraph{Protocol.}
Every test text is paired with $3$--$5$ atomic facts drafted from the \emph{original} only (never from
a reconstruction), so the facts cannot leak the condition. For each (original, reconstruction, facts)
item a rater labels each fact as preserved (P), distorted (D: present but with a wrong entity, number, or
relation), or missing (M), and flags whether the reconstruction asserts an \emph{invented} claim, {\em i.e.}, a
specific person, event, date, quantity, or place with no anchor anywhere in the original (a claim that
traces to a source sentence but was altered counts as D, not as invented). Raters see only the two texts and the facts. Strategy and retention rate are withheld. The grid covers $100$ BBC test articles
$\times$ \{Step, WordLen, WordFreq, LP\} $\times$ $\rkeep\in\{0.9,\dots,0.1\}$ (Gemini~2.0 Flash
reconstructions from the main experiments) and $100$ $512$-character BBC chunks $\times$ \{Entropy,
Entropy-LP, Hybrid-$\alpha{=}0.5$\} $\times$ $\rkeep\in\{0.9,0.7,0.5,0.3,0.1\}$ (Gemini~2.5 Flash
reconstructions; see the decoder version check below), $5{,}098$ items and about $25{,}000$ fact judgments per rater. Facts for $47$ texts are
the human-verified facts of the $54$-item human study. The rest were drafted by Gemini~2.5 Flash with
the same instructions. The raters are Claude Opus~5~\citep{anthropic2026opus5} and GPT-5.6~\citep{openai2026gpt56} (\texttt{gpt-5.6-sol}) with an identical
prompt. Nine items ($0.2\%$; eight safety refusals, one malformed output) are excluded.

\paragraph{Calibration.}
The raters are anchored on the blind human study of Appendix~\ref{app:human_eval}, which used the same
P/D/M labels and invented-claim flag on the same facts. On those $54$ items, per-fact agreement with each of the four human annotators is
Cohen's $\kappa=0.53$--$0.71$ for Claude and $0.54$--$0.59$ for GPT (three-way P/D/M), against
$0.40$--$0.59$ between the humans themselves. Binary agreement (P vs.\ not-P) is $0.63$--$0.75$ and
$0.51$--$0.69$. Both raters flag invented claims at the human scale ($5$ and $6$ of $54$ items; four items
were adjudicated as containing one). On the full grid the two raters agree with each other at
$\kappa=0.82$ (three-way) and $0.88$ (binary), with $86\%$ agreement on the invented-claim flag.

\begin{table*}[tbp]
\centering
\small
\setlength{\tabcolsep}{3.5pt}
\caption{Scaled fact-level evaluation on BBC News. Each cell reports \emph{fact preservation rate} (share
of facts judged P) / \emph{invented-claim rate} (share of items flagged) for the two LLM raters. Top block
(articles, Gemini~2.0 Flash decoder) and bottom block (512-character chunks, Gemini~2.5 Flash decoder)
are comparable within, not across, blocks. $n{=}100$ texts per cell. \textbf{Bold} = highest preservation rate per column within a block.}
\label{tab:fact_eval}
\begin{tabular}{l ccccc ccccc}
\toprule
& \multicolumn{5}{c}{\textbf{Claude Opus 5 rater}} & \multicolumn{5}{c}{\textbf{GPT-5.6 rater}} \\
\cmidrule(lr){2-6}\cmidrule(lr){7-11}
$\rkeep$ & 0.9 & 0.7 & 0.5 & 0.3 & 0.1 & 0.9 & 0.7 & 0.5 & 0.3 & 0.1 \\
\midrule
WordLen  & \textbf{1.00} / .00 & \textbf{.97} / .01 & \textbf{.64} / .15 & .15 / .28 & .00 / .31 & \textbf{1.00} / .00 & \textbf{.96} / .00 & \textbf{.56} / .17 & .11 / .34 & .00 / .61 \\
WordFreq & .93 / .00 & .80 / .03 & .56 / .10 & \textbf{.34} / .11 & \textbf{.12} / .06 & .91 / .00 & .74 / .03 & .48 / .15 & \textbf{.24} / .10 & \textbf{.07} / .09 \\
LP       & .91 / .00 & .79 / .02 & .53 / .09 & .24 / .09 & .09 / .04 & .90 / .00 & .73 / .02 & .46 / .03 & .12 / .13 & .03 / .01 \\
Step     & .90 / .00 & .42 / .19 & .01 / .34 & .00 / .36 & .00 / .29 & .90 / .00 & .41 / .16 & .08 / .53 & .00 / .73 & .00 / .71 \\
\midrule
Hybrid-$\alpha{=}0.5$ & \textbf{1.00} / .00 & \textbf{.92} / .00 & \textbf{.68} / .04 & \textbf{.33} / .06 & .03 / .31 & \textbf{.99} / .02 & \textbf{.88} / .04 & \textbf{.57} / .14 & \textbf{.25} / .23 & \textbf{.03} / .73 \\
Entropy-LP & .97 / .01 & .89 / .01 & .64 / .00 & .27 / .07 & \textbf{.04} / .30 & .96 / .06 & .82 / .06 & \textbf{.57} / .11 & .22 / .17 & \textbf{.03} / .68 \\
Entropy    & .98 / .01 & .88 / .03 & .61 / .02 & .25 / .08 & .03 / .31 & .95 / .07 & .81 / .04 & .54 / .08 & .21 / .27 & .02 / .69 \\
\bottomrule
\end{tabular}
\end{table*}

\paragraph{Findings.}
Table~\ref{tab:fact_eval} shows four patterns. (i)~The scaled protocol reproduces the human study:
for WordFreq and LP at $\rkeep=0.5/0.3/0.1$ the Claude rater gives preservation rates of
$0.56/0.34/0.12$ and $0.53/0.24/0.09$, against pooled human rates of $0.52/0.21/0.03$ on the same
strategies. (ii)~WordLen keeps facts best down to $\rkeep=0.5$ but collapses fastest below it, whereas
WordFreq and LP degrade gradually and are the only word-level strategies with non-trivial factual
recovery at $\rkeep\le0.2$. This is the fact-level counterpart of the regime shift in
Table~\ref{tab:unified}. (iii)~Step preserves essentially no facts from $\rkeep=0.5$ downward although
the BERTScore of those outputs stays above $0.80$, quantifying the leniency of BERTScore
for fluent but wrong text. (iv)~Invented claims are rare at $\rkeep\ge0.7$ ($\le4\%$ of items for WordLen, WordFreq and LP under both raters, $\le7\%$ for the semantic block under the GPT rater; Step is the exception at $16$--$19\%$) and reach $3$--$15\%$ for WordFreq and LP at $0.5$--$0.3$. At $\rkeep=0.1$ the two raters diverge (about $0.3$ vs.\ $0.7$ for WordLen and the semantic family), so we report the range and
treat that regime as a stress test. Within the semantic block, Hybrid-$\alpha{=}0.5$ preserves at least as many facts as Entropy-LP and Entropy at every rate for both raters, except $\rkeep=0.1$ under the Claude rater ($0.03$ against $0.04$, one item). Per-item judgments with the raters'
evidence spans, the item-to-condition mapping, and the rater prompt are released with the benchmark.

\paragraph{Decoder version check.}
The semantic block above and every analysis added after submission were decoded with Gemini~2.5 Flash,
while the word-level tables use Gemini~2.0 Flash. That model has been retired, and the 2.0 Flash outputs of
the semantic methods were not stored, so they cannot be re-scored or re-decoded. The word-level outputs
were stored: we re-decoded the WordFreq-3 skeletons of the $201$ BBC test articles with 2.5 Flash and scored
both versions with the same scorer (Table~\ref{tab:decoder_check}). Mean BERTScore F1 differs by $-0.009$ to
$+0.016$, on the order of the per-article standard deviations ($0.005$--$0.014$, Table~\ref{tab:wordfreq_detail}).
The newer decoder writes texts of about the original length even at low retention, where the older one wrote
shorter ones, which lifts its score at $\rkeep \le 0.3$. We therefore treat results across the two decoder
versions as comparable at the aggregate level, while keeping the original 2.0 Flash outputs as the
published record for the main tables.

\begin{table}[htbp]
\centering
\small
\setlength{\tabcolsep}{3pt}
\caption{Decoder version check on BBC News: mean BERTScore F1 of WordFreq-3 reconstructions from the same
skeletons ($201$ articles per cell), with the original Gemini~2.0 Flash outputs (the published values of
Table~\ref{tab:unified}) and a Gemini~2.5 Flash re-decoding, and their difference. The published outputs
used the word-level pipeline's prompts and the re-run the reconstruction prompt of the semantic methods, so
the check bounds decoder and prompt together.}
\label{tab:decoder_check}
\begin{tabular}{lccccc}
\toprule
Decoder & $0.9$ & $0.7$ & $0.5$ & $0.3$ & $0.1$ \\
\midrule
2.0 Flash & 0.9839 & 0.9645 & 0.9314 & 0.8948 & 0.8527 \\
2.5 Flash & 0.9752 & 0.9583 & 0.9338 & 0.9016 & 0.8689 \\
$\Delta$ & $-0.009$ & $-0.006$ & $+0.002$ & $+0.007$ & $+0.016$ \\
\bottomrule
\end{tabular}
\end{table}

\section{Downstream Task Evaluation}
\label{app:downstream}

BERTScore and the fact-level judgments of Appendix~\ref{app:fact_eval} both compare a reconstruction
against its original. Neither says whether the artifact is still \emph{usable}. We therefore evaluate
every deletion strategy at every retention rate on two standard text tasks, with two independent text
representations, scoring the skeleton and the reconstruction separately.

\paragraph{Protocol.}
The tasks are (i)~five-way topic classification, where a linear SVM is trained once on the ${\sim}2{,}010$
BBC \texttt{datasource} articles outside the test set and then applied unchanged to each variant, so the
classifier never sees degraded text at training time; and (ii)~summary-to-article retrieval, where the
test article's reference summary is the query, the $200$ variants at that strategy and rate form the
corpus, and we report Recall@$1$. Two backends are run independently: TF-IDF ($1$--$2$ grams) with a
linear SVM and cosine similarity, both from scikit-learn~\citep{pedregosa2011scikit}, and \texttt{all-MiniLM-L6-v2}~\citep{wang2020minilm} Sentence-BERT embeddings~\citep{reimers2019sentence} with the same classifier and metric. The embedding model truncates inputs to $256$ word-pieces, which $87\%$ of the
originals exceed, so its scores describe the article head, with every variant at a rate truncated alike. Word-level cells use the $200$--$201$ BBC test articles at all nine retention
rates (Table~\ref{tab:downstream}). Semantic and hybrid cells use the $100$ $512$-character chunks at
five rates (Table~\ref{tab:downstream_sem}).

\begin{table*}[tbp]
\centering
\small
\setlength{\tabcolsep}{5pt}
\caption{Downstream utility of word-level skeletons and their reconstructions on BBC News. Each cell is
\emph{topic accuracy} / \emph{retrieval Recall@1}, with $n{=}200$--$201$ test articles per cell. Original text
scores $.99$~/~$.99$ under the TF-IDF backend and $.96$~/~$.98$ under the neural backend. WordLen stores no
skeleton in the released result files, so it appears only in the reconstruction blocks. \textbf{Bold} = best
per column within a block, ranked on the sum of the two metrics. Ties are all marked.}
\label{tab:downstream}
\begin{tabular}{l ccccccccc}
\toprule
$\rkeep$ & 0.9 & 0.8 & 0.7 & 0.6 & 0.5 & 0.4 & 0.3 & 0.2 & 0.1 \\
\midrule
\multicolumn{10}{l}{\emph{TF-IDF backend: skeleton (no decoder)}} \\
Step         & .95 / .99 & .33 / .28 & .25 / .01 & .23 / .01 & .24 / .01 & .29 / .01 & .18 / .01 & .20 / .00 & .23 / .01 \\
WordFreq-3   & \textbf{.99 / .99} & \textbf{.99 / .99} & \textbf{.99 / .99} & \textbf{.98 / .99} & \textbf{.98 / .99} & \textbf{.98 / .99} & \textbf{.97 / .99} & \textbf{.93 / .99} & \textbf{.85 / .99} \\
WordFreq-6   & .98 / .99 & .98 / .99 & .98 / .99 & .96 / .99 & .97 / .99 & .95 / .99 & .88 / .98 & .79 / .93 & .64 / .65 \\
LP           & .94 / .98 & .53 / .06 & .54 / .05 & .56 / .06 & .52 / .20 & .53 / .41 & .56 / .51 & .61 / .56 & .53 / .61 \\
\addlinespace[2pt]
\multicolumn{10}{l}{\emph{TF-IDF backend: reconstruction}} \\
Step         & .98 / .99 & \textbf{.99 / .99} & .98 / .99 & .89 / .80 & .41 / .25 & .38 / .04 & .24 / .01 & .28 / .01 & .23 / .01 \\
WordLen      & \textbf{.99 / .99} & \textbf{.99 / .99} & \textbf{.99 / .99} & \textbf{.99 / .99} & .97 / .99 & .98 / .99 & .83 / .70 & .34 / .15 & .22 / .02 \\
WordFreq-3   & \textbf{.99 / .99} & \textbf{.99 / .99} & \textbf{.99 / .99} & \textbf{.99 / .99} & .97 / .99 & \textbf{.99 / .99} & \textbf{.97 / .99} & .95 / .99 & \textbf{.88 / .98} \\
WordFreq-6   & \textbf{.99 / .99} & \textbf{.99 / .99} & \textbf{.99 / .99} & .98 / .99 & \textbf{.98 / .99} & .98 / .99 & \textbf{.97 / .99} & \textbf{.96 / .99} & .88 / .96 \\
LP           & \textbf{.99 / .99} & .98 / .99 & .98 / .99 & .98 / .99 & .97 / .99 & .97 / .99 & .95 / .99 & .93 / .98 & .71 / .86 \\
\midrule
\multicolumn{10}{l}{\emph{Neural backend: skeleton (no decoder)}} \\
Step         & .92 / .90 & .41 / .23 & .26 / .02 & .32 / .03 & .28 / .03 & .29 / .03 & .23 / .01 & .20 / .01 & .22 / .01 \\
WordFreq-3   & \textbf{.96 / .99} & \textbf{.96 / .99} & \textbf{.97 / .99} & \textbf{.98 / .99} & \textbf{.97 / .99} & \textbf{.96 / .99} & \textbf{.95 / .98} & \textbf{.93 / .94} & \textbf{.89 / .87} \\
WordFreq-6   & \textbf{.97 / .98} & \textbf{.96 / .99} & \textbf{.97 / .99} & .96 / .98 & .96 / .96 & .94 / .95 & .91 / .90 & .88 / .82 & .75 / .63 \\
LP           & .94 / .95 & .68 / .66 & .69 / .59 & .61 / .53 & .58 / .40 & .57 / .35 & .62 / .41 & .66 / .45 & .59 / .44 \\
\addlinespace[2pt]
\multicolumn{10}{l}{\emph{Neural backend: reconstruction}} \\
Step         & .96 / .99 & .96 / .99 & .95 / .97 & .79 / .68 & .39 / .12 & .30 / .06 & .23 / .01 & .22 / .01 & .22 / .01 \\
WordLen      & .97 / .99 & .96 / .99 & .96 / .99 & .97 / .99 & .96 / .98 & .98 / .96 & .69 / .38 & .33 / .05 & .21 / .01 \\
WordFreq-3   & \textbf{.98 / .99} & .96 / .99 & \textbf{.98 / .99} & \textbf{.98 / .99} & \textbf{.97 / .99} & \textbf{.97 / .99} & \textbf{.98 / .99} & \textbf{.97 / .96} & \textbf{.89 / .83} \\
WordFreq-6   & .97 / .99 & \textbf{.97 / .99} & .97 / .99 & .97 / .99 & .96 / .99 & \textbf{.98 / .98} & .96 / .98 & .93 / .95 & \textbf{.90 / .82} \\
LP           & .96 / .99 & .96 / .99 & .96 / .99 & .97 / .97 & .96 / .97 & .94 / .94 & .93 / .92 & .91 / .84 & .75 / .62 \\
\bottomrule
\end{tabular}
\end{table*}

\begin{table*}[tbp]
\centering
\footnotesize
\setlength{\tabcolsep}{3pt}
\caption{Downstream utility of semantic, hybrid, and LLMLingua-2 skeletons and reconstructions on the
$100$ BBC chunks (Gemini~2.5 Flash decoder). Cells are \emph{single-class topic recall}: skeleton / reconstruction, {\em i.e.}, the share of
variants that the five-way classifier still labels with the chunk's true category. The chunks all come from
$18$ \texttt{politics} articles, so this is recall for one class rather than balanced accuracy, and
retrieval is not identifiable on this sample (see text). The classifier is trained on the datasource
articles \emph{excluding} those $18$ sources. Original chunks score $.93$ (TF-IDF) and $.95$ (neural).
LLMLingua-2 targets a compression rate rather than hitting it exactly: its realised character ratios are $.92$/$.76$/$.57$/$.34$/$.10$ across the five nominal rates, so at $\rkeep = 0.1$ it keeps \emph{fewer} characters than the entropy family ($.12$). \textbf{Bold} = best reconstruction per column. Ties are all marked.}
\label{tab:downstream_sem}
\begin{tabular}{l ccccc ccccc}
\toprule
& \multicolumn{5}{c}{\textbf{TF-IDF backend}} & \multicolumn{5}{c}{\textbf{Neural backend}} \\
\cmidrule(lr){2-6}\cmidrule(lr){7-11}
$\rkeep$ & 0.9 & 0.7 & 0.5 & 0.3 & 0.1 & 0.9 & 0.7 & 0.5 & 0.3 & 0.1 \\
\midrule
Entropy                    & .92 / .93 & .91 / \textbf{.94} & .87 / .91 & .71 / .84 & .35 / .58 & .95 / \textbf{.97} & .91 / .95 & .94 / .93 & .87 / .93 & .65 / .79 \\
Entropy-LP                 & .91 / \textbf{.94} & .91 / .90 & .87 / .89 & .75 / .82 & .40 / .56 & .94 / .96 & .92 / .93 & .92 / .92 & .89 / .93 & .70 / \textbf{.85} \\
Entropy-in-FreqBuckets     & .93 / \textbf{.94} & .89 / .92 & .74 / .84 & .58 / .72 & .38 / .47 & .96 / \textbf{.97} & .94 / .93 & .87 / .86 & .78 / .84 & .37 / .69 \\
Hybrid-$\alpha{=}0.3$      & .93 / \textbf{.94} & .90 / .92 & .86 / \textbf{.93} & .75 / .87 & .34 / .58 & .93 / .96 & .91 / \textbf{.96} & .91 / .94 & .89 / \textbf{.96} & .60 / .79 \\
Hybrid-$\alpha{=}0.5$      & .93 / .93 & .90 / .92 & .84 / .89 & .73 / .83 & .29 / .49 & .92 / .95 & .91 / .95 & .95 / .93 & .91 / .95 & .58 / .79 \\
Hybrid-$\alpha{=}0.7$      & .93 / .93 & .92 / .91 & .86 / .92 & .62 / .85 & .26 / .46 & .93 / .96 & .92 / .94 & .93 / .94 & .87 / .89 & .46 / .77 \\
LLMLingua-2                & .90 / .93 & .89 / .92 & .88 / .90 & .85 / \textbf{.89} & .55 / \textbf{.68} & .95 / .95 & .95 / .95 & .92 / \textbf{.95} & .93 / .92 & .70 / .84 \\
\bottomrule
\end{tabular}
\end{table*}

\paragraph{Reconstruction restores utility that deletion destroys.}
This is the clearest result in Table~\ref{tab:downstream}, and it is largest exactly where the skeleton is
least readable. Under TF-IDF at $\rkeep = 0.7$, the LP skeleton scores $.54$ accuracy and $.05$ Recall@1 (the right article is top-ranked for one query in twenty), while its reconstruction scores $.98$ and $.99$. The Step skeleton at the
same rate scores $.25$ and $.01$ against $.98$ and $.99$ after reconstruction. A skeleton that no lexical
retriever can match is therefore not an unusable artifact: the information needed to restore near-original
task performance is still present, and the decoder recovers it. This is the downstream counterpart of the
BERTScore and fact-level results, and it is the property that distinguishes deletion-for-reconstruction
from prompt compression, where the compressed string must itself carry the task.

\paragraph{WordFreq skeletons need no decoder at all.}
The WordFreq rows break the pattern: their skeletons stay usable on both tasks down to the most aggressive
rate, scoring $.85$ accuracy and $.99$ Recall@1 at $\rkeep = 0.1$ under TF-IDF ($.89$ / $.87$ neural),
within a few points of their own reconstructions. Because WordFreq deletes whole low-information words and
leaves high-information ones intact, the surviving tokens are exactly the ones both tasks rely on. Where a
consumer only needs to classify, route, or index text, a WordFreq skeleton can be consumed directly at a
tenth of the original length, with no decoder call, no latency, and no risk of reconstruction error, a regime the similarity metrics of Section~\ref{sec:main_results} cannot expose.

\paragraph{Character-level deletion destroys lexical matching immediately.}
Step's skeleton retrieval collapses from $.99$ to $.28$ between $\rkeep = 0.9$ and $0.8$ and reaches $.01$
by $0.7$, because character deletion leaves no intact word forms for TF-IDF to match. The neural backend
degrades along the same path ($.90 \to .23 \to .02$). Reconstruction masks this until $\rkeep = 0.6$ and
then fails too: below $0.5$, Step's reconstructions fall to the level of its skeletons ($.24$ / $.01$
against $.18$ / $.01$ at $\rkeep = 0.3$). The downstream floor therefore coincides with the fact-level floor of
Table~\ref{tab:fact_eval}, where Step preserves essentially no facts at those rates.

\paragraph{Semantic and hybrid strategies.}
The saved semantic, hybrid, and LLMLingua-2 artifacts are $512$-character chunks drawn from the head of the
BBC datasource, which places all $100$ of them in $18$ \texttt{politics} articles. Retrieval is therefore not
identifiable on this sample: with a per-chunk gold, several chunks compete for the same article summary and
even the \emph{original} chunks reach only $.18$ Recall@1, while with an article-level gold every variant
saturates at $1.00$. We report the classification task only, as single-class recall: the share of variants
that the five-way classifier, trained on the datasource articles excluding those $18$ sources, still labels \texttt{politics}. The original chunks score $.93$ (TF-IDF) and $.95$ (neural), so the ceiling is
genuine and there is real headroom below it. Table~\ref{tab:downstream_sem} shows the same qualitative
result as the word-level family: reconstruction exceeds the skeleton in $58$ of $70$ cells and is never lower by more than $.02$,
and the gap widens as retention falls, reaching $.26 \to .46$ (TF-IDF) and
$.46 \to .77$ (neural) for Hybrid-$\alpha{=}0.7$ at $\rkeep = 0.1$. One contrast deserves note.
LLMLingua-2~\citep{pan2024llmlingua} degrades far more gracefully than the rest of the family, and the gap
widens as retention falls: at $\rkeep = 0.1$ its skeleton scores $.55$ against $.26$--$.40$ for the entropy
and hybrid strategies, and its reconstruction $.68$ against $.46$--$.58$. It achieves this while keeping
\emph{fewer} characters than they do ($.10$ of the original against $.12$ for Entropy), so the advantage
lies in which tokens it keeps rather than how many. That is the expected signature of a prompt compressor:
LLMLingua-2 is trained to preserve the tokens a downstream consumer needs, which is precisely what this
task rewards. It is a useful reminder that skeleton task utility and reconstruction fidelity are distinct
objectives, so a method that leads on one need not lead on the other.

\paragraph{Backend agreement.}
The two backends agree on every collapse point and on every ordering of reconstructions separated by more
than $0.02$, so the conclusions are not an artifact of one representation. They differ in level and in where they are forgiving. MiniLM scores
the originals lower ($.96$ / $.98$ against $.99$ / $.99$) because a $384$-dimensional sentence embedding is
weaker than TF-IDF $n$-grams on a five-way news task. Embeddings are markedly more tolerant of broken word forms (the LP skeleton at $\rkeep = 0.7$ retrieves at $.59$ under MiniLM against $.05$ under TF-IDF), while TF-IDF is the more optimistic of the two on heavily deleted WordFreq skeletons, whose surviving rare
words are strong exact-match keys ($.99$ against $.87$ Recall@1 at $\rkeep = 0.1$). Full nine-rate results
for both backends, including macro-F1 and Recall@5, are in the released evaluation artifact.

\section{LLMLingua-2 as an Encoder}
\label{app:llmlingua2}

Prompt compression is the closest external line of work, so we test whether a compressor built for
that objective transfers to ours. Original LLMLingua~\citep{jiang2023llmlingua} cannot be dropped in
unchanged: its perplexity-based selection runs a billion-scale LM aligned to the target model at encode
time, contradicting the weak-sender setting this benchmark studies (our encoders use at most a $124$M
surrogate), and its budget controller targets a context window rather than a retention rate. We
therefore run LLMLingua-2~\citep{pan2024llmlingua}, the classifier-based, rate-targeted successor, as a
drop-in encoder on the same $100$ BBC chunks with the same reconstruction prompt and decoder as the
semantic family, at all five retention rates (Table~\ref{tab:llmlingua2}). Three findings follow.

\paragraph{It does not dominate on the reconstruction objective.}
In Table~\ref{tab:semantic_baselines}, LLMLingua-2 trails Entropy-LP by
$0.004$--$0.008$ BERTScore from $\rkeep = 0.9$ to $0.5$, ties it at $0.3$ ($0.9121$ against $0.9115$), and
leads the table at $0.1$ ($0.8805$ against $0.8752$ for WordFreq), where every method is close to its floor. Its rate
control is also approximate rather than exact, retaining $0.917/0.756/0.568/0.341/0.097$ of the
characters at the five nominal rates.

\paragraph{It does not optimise recoverability.}
Scored as-is, with no reconstruction (the way prompt compression itself would consume the output), it trails its own reconstructed version by $2.1$, $3.3$, $2.6$ and $1.9$ BERTScore points at
$\rkeep = 0.7$, $0.5$, $0.3$ and $0.1$. The one exception is instructive: at $\rkeep = 0.9$ the raw
skeleton scores $0.0007$ \emph{above} its reconstruction, because at $92\%$ retention the artifact is
already almost the original and the decoder can only add noise. Reconstruction pays for itself in the
middle of the range, not at its ends.

\paragraph{But it preserves entities better wherever compression is substantial.}
From $\rkeep = 0.5$ downward LLMLingua-2 leads every strategy in our semantic family on both entity
recall and unsupported-entity rate, and the margin widens as retention falls: at $\rkeep = 0.1$ it
recalls $0.307$ of the entities against $0.177$ for the best of the family. At $\rkeep = 0.9$ it is
behind (Hybrid-$\alpha{=}0.5$ recalls $0.988$ against its $0.974$), so this is a low-retention
advantage rather than a blanket one. A token classifier trained to keep what a downstream consumer
needs keeps named entities, and entities are what our factual metrics reward, which is why it leads
the downstream task evaluation of Appendix~\ref{app:downstream} at aggressive rates while merely
matching on BERTScore. Skeleton task utility, factual preservation and reconstruction fidelity are
three different objectives, and no single encoder led on all three.

\begin{table}[htbp]
\centering
\small
\setlength{\tabcolsep}{3.5pt}
\caption{LLMLingua-2 inside our pipeline on $100$ BBC chunks (Gemini~2.5 Flash decoder). Recon.\ =
BERTScore F1 of the reconstruction; Skel.\ = the compressed output scored directly against the original,
with no reconstruction, as prompt compression itself would use it; $\hat{r}$ = realised character ratio
against the nominal $\rkeep$; Ent.\ = entity recall; Uns.\ = unsupported-entity rate.}
\label{tab:llmlingua2}
\begin{tabular}{lccccc}
\toprule
$\rkeep$ & $\hat{r}$ & Recon. & Skel. & Ent. & Uns. \\
\midrule
0.9 & 0.917 & 0.9857 & 0.9864 & 0.974 & 0.043 \\
0.7 & 0.756 & 0.9709 & 0.9503 & 0.964 & 0.052 \\
0.5 & 0.568 & 0.9462 & 0.9134 & 0.856 & 0.123 \\
0.3 & 0.341 & 0.9121 & 0.8862 & 0.643 & 0.260 \\
0.1 & 0.097 & 0.8805 & 0.8614 & 0.307 & 0.478 \\
\bottomrule
\end{tabular}
\end{table}

\section{Encoder and Decoder Levers}
\label{app:levers}

Table~\ref{tab:finetune} changes the decoder while Tables~\ref{tab:unified}
and~\ref{tab:semantic_baselines} change the encoder, which leaves open how the two compare. We measure them
against each other on the same axis. The \emph{decoder lever} is the paired BERTScore gain of the
QLoRA-tuned Llama-3.2-3B-Instruct over the same base model on identical skeletons, about $800$
paired chunks per rate (Wilcoxon $p < 10^{-99}$ at every rate), the pairs behind Table~\ref{tab:finetune},
whose ZS-to-SFT differences these gains reproduce within $0.001$. The \emph{encoder lever} is the best-worst
spread across the word-level deletion strategies at a fixed decoder, which is the range a practitioner
moves through by changing the deletion rule alone.

The two are comparable in magnitude, and which one dominates depends on the operating point
(Table~\ref{tab:levers}). The encoder lever is larger at $\rkeep = 0.7$ and $0.1$, the decoder lever
at $0.9$, $0.5$ and $0.3$, with the two within $0.005$ only at $0.3$. This matters for reading
Table~\ref{tab:finetune}: decoder adaptation is not a substitute for choosing the deletion rule, and at
the lowest rate, where our strategy claims are strongest, the rule moves BERTScore almost twice as far as
the decoder does. The character-level Step baselines are excluded from the encoder spread because including a deliberately naive baseline would inflate it. The spread covers the word-level strategies on the stored Gemini~2.0 Flash reconstructions, scored as in Table~\ref{tab:unified}. The figures below are therefore the conservative ones.

\begin{table}[htbp]
\centering
\small
\setlength{\tabcolsep}{4pt}
\caption{The two levers on the same axis (BERTScore F1). \emph{Decoder}: paired gain of QLoRA over the base
Llama decoder on identical skeletons, about $800$ paired chunks per rate. \emph{Encoder}: best-worst spread
across the word-level deletion strategies at a fixed decoder. The larger lever per rate is in
\textbf{bold}.}
\label{tab:levers}
\begin{tabular}{lcc}
\toprule
$\rkeep$ & Decoder & Encoder \\
\midrule
0.9 & \textbf{0.011} & 0.005 \\
0.7 & 0.014 & \textbf{0.023} \\
0.5 & \textbf{0.056} & 0.035 \\
0.3 & \textbf{0.050} & 0.046 \\
0.1 & 0.039 & \textbf{0.069} \\
\bottomrule
\end{tabular}
\end{table}

\section{Bucket Granularity Ablation}
\label{app:ablations}

WordFreq partitions the vocabulary into frequency buckets and spreads the deletion quota across them
(Section~\ref{sec:wordfreq}). The LP variant uses a six-bucket taxonomy. Table~\ref{tab:bucket_ablation}
asks whether the finer split helps the static WordFreq rule itself, on the BBC News benchmark
($n{=}200$ chunks, Gemini~2.0 Flash decoder, identical pipeline otherwise).

\begin{table}[htbp]
\centering
\small
\setlength{\tabcolsep}{5pt}
\caption{Ablation: 3-class vs.\ 6-class frequency bucketing for WordFreq deletion on BBC News
(BERTScore F1, mean over $200$ chunks; CER in parentheses). \textbf{Bold} = better F1 per row.}
\label{tab:bucket_ablation}
\begin{tabular}{lcc}
\toprule
$\rkeep$ & 3-Class & 6-Class \\
\midrule
0.9 & 0.9839 (0.104) & \textbf{0.9850} (0.094) \\
0.8 & 0.9754 (0.160) & \textbf{0.9768} (0.151) \\
0.7 & \textbf{0.9645} (0.227) & 0.9636 (0.224) \\
0.6 & \textbf{0.9479} (0.307) & 0.9441 (0.316) \\
0.5 & \textbf{0.9314} (0.400) & 0.9236 (0.434) \\
0.4 & \textbf{0.9137} (0.501) & 0.9001 (0.565) \\
0.3 & \textbf{0.8948} (0.615) & 0.8779 (0.690) \\
0.2 & \textbf{0.8737} (0.733) & 0.8612 (0.796) \\
0.1 & \textbf{0.8527} (0.851) & 0.8422 (0.896) \\
\bottomrule
\end{tabular}
\end{table}

The two granularities are indistinguishable at mild compression (six classes are ahead by
$0.001$ at $\rkeep\ge0.8$), but from $\rkeep=0.7$ downward the coarser three-class split is
consistently better, by $0.8$ points at $\rkeep=0.5$, $1.7$ points at $0.3$, and $1.1$ points at
$0.1$, with lower character error rate at every rate $\le0.6$. Finer buckets spread the same deletion
budget over more, smaller classes, so once the budget is large a proportionally larger share of it
falls on mid-frequency content words that carry most of the recoverable meaning, whereas the
three-class rule concentrates deletions on the high-frequency class for longer. We therefore use the
three-class variant as the default WordFreq configuration in the main experiments and reserve the
six-bucket taxonomy for the LP allocation, where the buckets are weighted rather than treated equally.

\section{Extreme Compression: $\rkeep = 0.05$}
\label{app:r005}
\begin{table*}[tbp]
\centering
\small
\setlength{\tabcolsep}{4pt}
\caption{Beyond the paper's lowest rate: $\rkeep = 0.05$ against $\rkeep = 0.1$ under one decoder (Gemini~2.5 Flash). Each cell is $\rkeep{=}0.1$~/~$0.05$. \emph{Length} is the
reconstruction's character count over the original's. WordFreq-3 uses the $201$ BBC test articles of Table~\ref{tab:unified}. The semantic rows are the $100$ chunks of
Table~\ref{tab:semantic_baselines}, with $n{=}99$ where the API safety filter refused one three-word
skeleton. \textbf{Bold} = best per column within the semantic block.}
\label{tab:r005}
\begin{tabular}{l cc cc cc cc cc}
\toprule
& \multicolumn{2}{c}{\textbf{BERTScore}} & \multicolumn{2}{c}{\textbf{ROUGE-L}} & \multicolumn{2}{c}{\textbf{Entity recall}} & \multicolumn{2}{c}{\textbf{Unsupported}} & \multicolumn{2}{c}{\textbf{Length}} \\
\cmidrule(lr){2-3}\cmidrule(lr){4-5}\cmidrule(lr){6-7}\cmidrule(lr){8-9}\cmidrule(lr){10-11}
$\rkeep$ & 0.1 & 0.05 & 0.1 & 0.05 & 0.1 & 0.05 & 0.1 & 0.05 & 0.1 & 0.05 \\
\midrule
WordFreq-3 ($n{=}201$)     & .854 & .841 & .197 & .125 & .281 & .203 & .206 & .240 & 0.19 & 0.11 \\
\midrule
Entropy                    & \textbf{.874} & .863 & .154 & .120 & \textbf{.177} & .083 & .538 & \textbf{.748} & 1.28 & 1.38 \\
Entropy-LP                 & \textbf{.874} & \textbf{.865} & \textbf{.163} & \textbf{.126} & .177 & \textbf{.103} & \textbf{.533} & .769 & 1.22 & 1.28 \\
Entropy-in-FreqBuckets     & .869 & .862 & .159 & .123 & .072 & .048 & .646 & .767 & 1.11 & 1.33 \\
Hybrid-$\alpha{=}0.3$      & .872 & .863 & .146 & .119 & .173 & .084 & .577 & .777 & 1.24 & 1.32 \\
Hybrid-$\alpha{=}0.5$      & .872 & .862 & .144 & .114 & .171 & .072 & .583 & .795 & 1.28 & 1.39 \\
Hybrid-$\alpha{=}0.7$      & .871 & .862 & .132 & .112 & .127 & .065 & .678 & .780 & 1.21 & 1.45 \\
\bottomrule
\end{tabular}
\end{table*}

Table~\ref{tab:r005} adds an $\rkeep = 0.05$ point, at which a $512$-character chunk keeps about
$30$ characters. Both rates were decoded with Gemini~2.5 Flash. For WordFreq-3 the word-level pipeline reproduces the published $\rkeep = 0.1$ cells within
$0.002$ BERTScore (Table~\ref{tab:unified}) and $0.013$ entity recall (Table~\ref{tab:ner_recon}), so the decoder change does not affect the comparison.

The point tests the prediction that a sparser skeleton yields fluent but entity-free text, and the
prediction holds. Halving the skeleton costs the semantic methods about $0.01$ BERTScore and $0.03$
ROUGE-L, while entity recall halves to $0.05$--$0.10$ and the unsupported-entity rate reaches
$0.75$--$0.80$: three entities in four in the output have no counterpart in the source. Those
reconstructions are $1.3$--$1.45\times$ the original length, so the decoder is not truncating but
composing. WordFreq-3 degrades the same way but from a higher floor: its decoder restores only
what the skeleton supports, returning $0.19$ and then $0.11$ of the original length, and at
$\rkeep = 0.05$ it still retains $0.20$ of the entities with an unsupported rate of $0.24$, the lowest in
the table, because the entities it keeps ride in the skeleton rather than being regenerated
(Section~\ref{sec:analysis}). Nothing below $\rkeep = 0.1$
changes the paper's recommendation: the regime is a stress test, and the metric that reports it as one is
entity recall, not BERTScore.

\section{Lossless Compression Baselines}
\label{app:lossless_baselines}

\begin{table*}[tbp]
\centering
\small
\setlength{\tabcolsep}{4pt}
\caption{Three-tier storage accounting on BBC News, as a fraction of the original bytes (lower is better).
Each cell is \emph{tier 1} (the skeleton alone) / \emph{tier 2} (the skeleton after zlib). For reference,
the same codecs applied to the \emph{original} text give $0.513$ (zlib), $0.534$ (bzip2) and $0.579$ (LZMA).
Tier 2 uses zlib because zlib is the strongest of the three on this data on \emph{both} sides of the
comparison: on the original above, and composed with the skeleton in every cell below, where bzip2 and
LZMA are worse by up to $0.07$ at these $512$-character chunk sizes. The comparison is therefore
like-for-like and uses each side's best codec.
Tier 2 is the storage figure of merit: it beats the lossless baseline at every rate, including at
$\rkeep = 0.9$, where the skeleton alone does not. Tier 3 (encoder latency, decoder weights, prompt bytes, reconstruction
latency) is a system cost, not a byte count, and is discussed in the text. Character-level Step is omitted.}
\label{tab:tiers}
\begin{tabular}{lccccc}
\toprule
$\rkeep$ & 0.9 & 0.7 & 0.5 & 0.3 & 0.1 \\
\midrule
WordLen     & .886 / .498 & .699 / .411 & .499 / .308 & .299 / .203 & .099 / .076 \\
WordFreq-3  & .902 / .472 & .711 / .391 & .519 / .304 & .327 / .209 & .137 / .104 \\
WordFreq-6  & .891 / .476 & .688 / .390 & .490 / .293 & .290 / .186 & .090 / .072 \\
LP          & .888 / .496 & .690 / .394 & .491 / .289 & .293 / .190 & .094 / .075 \\
\bottomrule
\end{tabular}
\end{table*}

Classical lossless codecs operate on the byte stream without semantic interpretation.
On the BBC News test set ($n{=}200$ chunks of 512 characters),
zlib achieves ${\sim}1.9\times$ compression ($\rkeep \approx 0.52$) and
LZMA ${\sim}1.7\times$ ($\rkeep \approx 0.58$), both with zero distortion.
LZMA's slightly lower ratio at this chunk size is due to its higher header overhead.

Section~\ref{sec:dataset} reports tier 1, the skeleton as a fraction $\rkeep$ of the original characters.
Two further tiers complete the accounting. \emph{Tier 2} composes that skeleton with a lossless codec. The
composition is free: the codec decodes exactly, so the decoder receives the same skeleton and returns the
same reconstruction, and the only change is the number of bytes at rest. At $\rkeep = 0.5$ a WordFreq-6
skeleton occupies $0.293$ of the original bytes after zlib, against $0.513$ for zlib applied to the original. Table~\ref{tab:tiers} gives both tiers for every strategy and rate. \emph{Tier 3} is the system cost: encoder latency (Table~\ref{tab:latency}; ${\sim}1$\,ms per chunk for
WordFreq), decoder weights ($6.4$\,GB in fp16, a $195$\,MB adapter, or zero local bytes for an API
decoder), $200$--$400$ prompt bytes, and $1$--$3$\,s of latency per reconstructed chunk. The weights are
one-time and amortized. The prompt bytes and decoding latency are paid per recovery, and only when recovery is
requested. Two consumption paths therefore differ in what tier 3 costs and what the pipeline saves.
\emph{Reconstruct-first} pays the $1$--$3$\,s per chunk and saves bytes at rest and in transit only: the
consumer receives a text of roughly the original length $L$, so there is no downstream inference or memory saving.
\emph{Skeleton-direct} pays nothing at tier 3 and shortens the consumer's input to about $\rkeep L$. Appendix~\ref{app:downstream} shows which skeletons remain usable that way (WordFreq down to
$\rkeep = 0.1$, character-level Step not at all). Reporting
$1/\rkeep$ alone is therefore incomplete in both directions: it understates what the pipeline stores, since tier
2 is free, and overstates the end-to-end benefit, since tier 3 is not.

These numbers also settle which operating points are worth using. The storage advantage does not require
aggressive compression: at $\rkeep = 0.9$--$0.8$ the pipeline already stores $0.43$--$0.50$ of the original
bytes, below zlib's $0.513$ on the original text, while preserving $89$--$95\%$ of named entities with
$3$--$6\%$ unsupported. Table~\ref{tab:regimes} maps the three regimes this produces. Mild retention is the
archival and agent-memory setting: near-original fidelity at a storage cost already better than lossless.
Moderate retention buys two to three times the lossless ratio and stays usable for gist preservation, logs
and previews, provided entity-critical fields are stored separately. Aggressive retention gives the largest
ratios but loses most factual anchors, and is best read as a stress test of decoder robustness (followed to $\rkeep = 0.05$ in Appendix~\ref{app:r005}) rather than
an operating point. The comparison is meaningful because of how the two degrade. A lossless codec cannot be pushed below its
own ratio, and forcing a byte-level method past that point returns corrupted bytes. The skeleton instead
degrades into text that stays coherent and readable. That is what makes the trade worth taking when the
consumer is a reader or a semantic processor rather than one requiring exact bit recovery. Lossless coding
is therefore a complement to our approach rather than a competitor, and the comparison does not rest on the
regime in which reconstruction fails.

\begin{table*}[tbp]
\centering
\small
\setlength{\tabcolsep}{5pt}
\caption{Operating regimes on BBC News. \emph{Stored} is tier 2, the skeleton after zlib, as a fraction of the original bytes, and zlib applied to the original text gives $0.513$. Bands span WordFreq and LP, the two
strategies we recommend, over the rates in each regime. Entity recall and the unsupported-entity rate are
from Table~\ref{tab:ner_recon} and the fact-level rates from Appendix~\ref{app:fact_eval}.}
\label{tab:regimes}
\begin{tabular}{llcccl}
\toprule
Regime & $\rkeep$ & Stored & Entity recall & Unsupported & Suitable for \\
\midrule
Mild       & $0.9$--$0.8$ & $0.43$--$0.50$ & $0.89$--$0.95$ & $0.03$--$0.06$ & archival, agent memory \\
Moderate   & $0.7$--$0.5$ & $0.29$--$0.39$ & $0.65$--$0.88$ & $0.07$--$0.15$ & gist preservation: logs, previews \\
Aggressive & $0.3$--$0.1$ & $0.08$--$0.21$ & $0.19$--$0.54$ & $0.16$--$0.26$ & stress test, not an operating point \\
\bottomrule
\end{tabular}
\end{table*}

\section{Choosing a Strategy}
\label{app:choosing}

Appendix~\ref{app:lossless_baselines} maps retention rates to what they cost and buy.
This section addresses the complementary question a practitioner faces on a new domain: which deletion
rule to start from. The measurements in this paper support a short default.

\emph{Start with WordFreq.} It needs no model at the encoder (${\sim}1$\,ms per chunk against
${\sim}15$\,ms GPU or ${\sim}130$\,ms CPU for anything surprisal-based, Table~\ref{tab:latency}), and it is
the most robust choice once retention is low: at $\rkeep = 0.3$ and $0.1$ it has both the highest entity
recall of the word-level strategies ($0.544$ and $0.268$) and the lowest unsupported-entity rate ($0.158$
and $0.237$), and it leads on BERTScore at both rates and on ROUGE-L at $\rkeep = 0.1$ (Table~\ref{tab:unified}). It is also the
only strategy whose skeletons remain directly usable without a decoder, at $0.85$ topic accuracy and $0.99$
retrieval Recall@1 even at $\rkeep = 0.1$ (Appendix~\ref{app:downstream}), which removes the decoder cost
entirely when the consumer is a classifier or an index rather than a reader.

\emph{Deviate for a reason, and only in the middle of the range.} At $\rkeep \approx 0.5$ the ranking
depends on which fidelity is wanted, and the metrics do not agree. LP is best on both similarity metrics and
preserves more facts than WordFreq-6 in the human study ($58\%$ against $47\%$), but WordLen has the highest
entity recall at that rate ($0.760$ against $0.722$ for WordFreq-3 and $0.653$ for LP). Surprisal-based and
hybrid encoders are worth their cost only here as well, and only when an encoder-side model is available. Below $\rkeep = 0.3$ they no longer beat WordFreq (Table~\ref{tab:semantic_baselines}).

\emph{Select on more than one metric.} The ranking is domain-dependent, and the cross-domain results that
follow (Appendices~\ref{app:wikipedia_eval}--\ref{app:chinese_eval}) show it shifting with register and
information density, so the default above is a starting point rather than a conclusion. On a held-out sample
from a new domain, sweep the rates and compare strategies on a lexical metric and a factual one alongside
BERTScore, not on BERTScore alone: it saturates
near $0.81$ and will rank strategies that have lost most of their content as acceptable
(Appendix~\ref{app:detailed_metrics}). Where the reconstruction will be consumed by a task rather than read,
measure that task directly, since the ordering there is not the same as the ordering on similarity. The
released harness runs these sweeps.

\section{Cross-Domain Evaluation: Wikipedia}
\label{app:wikipedia_eval}

To assess cross-domain generalizability, we evaluate our deletion strategies on the
WikiText corpus~\citep{merity2016wikitext}, taken from the Salesforce/wikitext release~\citep{salesforce2022wikitext} on HuggingFace,
a collection of encyclopedic Wikipedia articles spanning diverse factual domains.
This domain differs substantially from BBC News in register (encyclopedic vs.\ journalistic),
sentence structure (longer, more complex), and vocabulary (higher technical density).
We use the same evaluation protocol as the main experiments.

\textbf{Results.}
Table~\ref{tab:wiki_bertscore} reports BERTScore F1 across all deletion strategies
at representative retention rates.
The Entropy, Entropy-LP, Entropy-in-FreqBuckets, and Hybrid rows use the same
GPT-2 surprisal pipeline and Gemini 2.0 Flash decoder as the BBC News experiments
(Section~\ref{sec:new_baselines}).
Among the zero-shot methods the semantic family leads at every rate: pure Entropy at
$\rkeep = 0.9$ ($0.9820$), Hybrid-$\alpha{=}0.3$ at $0.7$ ($0.9718$), and Entropy-LP from $0.5$
downward ($0.9423$, $0.8927$, $0.8445$). The frequency family stays close where it is strongest:
WordLen trails by $0.001$ at $\rkeep = 0.7$, and the better of LP and WordFreq-3 trails Entropy-LP by
$0.011$ at $\rkeep = 0.3$ and $0.005$ at $0.1$. Encyclopedic text holds many terms that are globally rare
but locally predictable. A frequency table keeps such terms even when the context already predicts
them, whereas surprisal deletes them and saves the budget for tokens the decoder could not guess,
which fits the lead of surprisal-based allocation. Direct rank interpolation adds little: the
Hybrid rows track pure Entropy and fall behind Entropy-LP once the skeleton becomes sparse.

\begin{table*}[tbp]
\centering
\small
\caption{BERTScore F1 on Wikipedia (Salesforce/wikitext).
Entropy and Hybrid rows use GPT-2 surprisal (same pipeline as BBC News experiments).
Llama base and Llama SFT report the local Llama-3.2-3B-Instruct decoder before and after QLoRA fine-tuning; at each rate both use the deletion strategy whose fine-tuned decoder performs best (WordLen at $\rkeep \ge 0.5$, WordFreq-3 at $\rkeep \le 0.3$).
\textbf{Bold} = best per column within each zero-shot / local-decoder block.}
\label{tab:wiki_bertscore}
\begin{tabular}{lccccc}
\toprule
\textbf{Method} & $\rkeep{=}0.9$ & $\rkeep{=}0.7$ & $\rkeep{=}0.5$ & $\rkeep{=}0.3$ & $\rkeep{=}0.1$ \\
\midrule
Step                    & $0.9715$ & $0.9473$ & $0.8305$ & $0.8131$ & $0.8032$ \\
WordLen                 & $0.9724$ & $0.9705$ & $0.9334$ & $0.8385$ & $0.8010$ \\
WordFreq-3               & $0.9639$ & $0.9432$ & $0.9162$ & $0.8812$ & $0.8398$ \\
WordFreq-6               & $0.9666$ & $0.9430$ & $0.9069$ & $0.8644$ & $0.8282$ \\
LP                  & $0.9695$ & $0.9577$ & $0.9273$ & $0.8816$ & $0.8342$ \\
Entropy                 & $\mathbf{0.9820}$ & $0.9715$ & $0.9414$ & $0.8908$ & $0.8438$ \\
Entropy-LP              & $0.9810$ & $0.9716$ & $\mathbf{0.9423}$ & $\mathbf{0.8927}$ & $\mathbf{0.8445}$ \\
Entropy-in-FreqBuckets  & $0.9770$ & $0.9556$ & $0.9145$ & $0.8765$ & $0.8365$ \\
Hybrid-$\alpha{=}0.3$   & $0.9801$ & $\mathbf{0.9718}$ & $0.9403$ & $0.8899$ & $0.8426$ \\
Hybrid-$\alpha{=}0.5$   & $0.9803$ & $0.9708$ & $0.9329$ & $0.8860$ & $0.8372$ \\
Hybrid-$\alpha{=}0.7$   & $0.9790$ & $0.9706$ & $0.9278$ & $0.8772$ & $0.8290$ \\
\midrule
Llama base              & $0.9708$ & $0.9344$ & $0.8288$ & $0.8627$ & $0.8313$ \\
Llama SFT               & $\mathbf{0.9862}$ & $\mathbf{0.9797}$ & $\mathbf{0.9514}$ & $\mathbf{0.9093}$ & $\mathbf{0.8756}$ \\
\bottomrule
\end{tabular}
\end{table*}

\textbf{Key findings:}
\begin{enumerate}
    \item \textbf{Frequency methods transfer but no longer lead.} WordFreq-3 and LP need no domain-specific recalibration and LP stays within $0.015$ of the best zero-shot method from $\rkeep = 0.5$ downward, but the semantic family is ahead at every rate.
    \item \textbf{Surprisal-based allocation leads throughout.} Entropy is strongest at $\rkeep=0.9$, Hybrid-$0.3$ at $\rkeep=0.7$, and Entropy-LP from $\rkeep=0.5$ downward, so on encyclopedic text budget allocation over surprisal buckets is the most reliable zero-shot choice.
    \item \textbf{Hybrid gains are small.} Direct frequency--surprisal interpolation tracks pure Entropy and weakens as the frequency weight grows (Hybrid-$0.7$ is the weakest hybrid from $\rkeep = 0.5$ downward), likely because the frequency component keeps terms that are rare globally but predictable locally.
    \item \textbf{Fine-tuning helps at every rate.} Llama SFT improves over Llama base by $0.015$--$0.123$ and leads every zero-shot method at every rate, including $\rkeep = 0.1$ ($0.8756$ against $0.8445$ for Entropy-LP).
    \item \textbf{Step remains the weakest deletion family once compression is non-trivial.} Its deterioration mirrors the main-paper pattern, confirming that uniform deletion does not transfer well across domains.
\end{enumerate}

\section{Cross-Domain Evaluation: Reddit}
\label{app:reddit_eval}

To further assess cross-domain robustness, we evaluate our deletion strategies on a
Reddit comments dataset~\citep{mentionbroker2023reddit} (parent comments from the Reddit Training Dataset; $n{=}200$ held-out chunks
of $\le 512$ characters for the word-level strategies and $n{=}100$ for the semantic and hybrid strategies), a conversational register that differs substantially from BBC News
(journalistic) and Wikipedia (encyclopedic) in both sentence length and vocabulary distribution.
We use the same evaluation protocol and Gemini 2.0 Flash zero-shot decoder as in the main experiments.

\textbf{Results.}
Table~\ref{tab:reddit_bertscore} reports BERTScore F1 for the frequency-family,
entropy-based, and hybrid deletion strategies, along with the fine-tuned Llama-3.2-3B-Instruct (SFT),
across five representative retention rates.

\begin{table*}[tbp]
\centering
\small
\caption{BERTScore F1 on Reddit conversational text.
$n{=}200$ for the word-level strategies and $n{=}100$ for the semantic and hybrid strategies.
WordFreq-3 (char.) applies character-level retention targeting instead of token-level.
Entropy and Hybrid rows use the same GPT-2 surprisal + Gemini 2.0 Flash pipeline as in the
BBC experiments. Llama base and Llama SFT report the local Llama-3.2-3B-Instruct decoder before and after QLoRA fine-tuning, evaluated on the same single-strategy Reddit deletion inputs used for local fine-tuning.
\textbf{Bold} = best per column within each zero-shot / local-decoder block.}
\label{tab:reddit_bertscore}
\begin{tabular}{lccccc}
\toprule
\textbf{Method} & $\rkeep{=}0.9$ & $\rkeep{=}0.7$ & $\rkeep{=}0.5$ & $\rkeep{=}0.3$ & $\rkeep{=}0.1$ \\
\midrule
Step              & $0.9928$ & $0.9632$ & $0.8465$ & $0.8284$ & $0.8161$ \\
WordLen           & $0.9926$ & $0.9913$ & $0.9378$ & $0.8572$ & $0.8194$ \\
WordFreq-3         & $0.9836$ & $0.9591$ & $0.9231$ & $0.8841$ & $0.8439$ \\
WordFreq-3 (char.) & $\mathbf{0.9937}$ & $\mathbf{0.9924}$ & $0.9392$ & $0.8573$ & $0.8184$ \\
WordFreq-6         & $0.9854$ & $0.9568$ & $0.9141$ & $0.8665$ & $0.8346$ \\
LP            & $0.9911$ & $0.9766$ & $0.9370$ & $0.8848$ & $0.8375$ \\
Entropy           & $0.9924$ & $0.9762$ & $0.9429$ & $0.8936$ & $0.8503$ \\
Entropy-LP        & $0.9924$ & $0.9781$ & $\mathbf{0.9432}$ & $0.8961$ & $\mathbf{0.8512}$ \\
Entropy-in-FreqBuckets & $0.9887$ & $0.9568$ & $0.9212$ & $0.8844$ & $0.8495$ \\
Hybrid-$0.3$      & $0.9934$ & $0.9783$ & $0.9425$ & $\mathbf{0.8969}$ & $0.8505$ \\
Hybrid-$0.5$      & $0.9934$ & $0.9782$ & $0.9393$ & $0.8951$ & $0.8491$ \\
Hybrid-$0.7$      & $0.9931$ & $0.9796$ & $0.9363$ & $0.8849$ & $0.8465$ \\
\midrule
Llama base      & $0.9780$ & $0.9565$ & $0.8505$ & $0.8668$ & $0.8371$ \\
Llama SFT       & $\mathbf{0.9946}$ & $\mathbf{0.9927}$ & $\mathbf{0.9590}$ & $\mathbf{0.9019}$ & $\mathbf{0.8525}$ \\
\bottomrule
\end{tabular}
\end{table*}

\textbf{Key findings:}
\begin{enumerate}
    \item \textbf{Reddit is unusually easy at high retention.} Many methods cluster near the top of the table at $\rkeep=0.9$, reflecting the shorter, more repetitive structure of conversational text.
    \item \textbf{Frequency methods still generalize, but semantic methods are competitive.} WordFreq-3 and LP remain strong, yet the best semantic and hybrid methods match or exceed them at most rates.
    \item \textbf{Conversational text benefits from contextual signals.} As on Wikipedia, Reddit often favors surprisal-based ranking or hybridization, suggesting that local contextual redundancy matters more here than corpus-level frequency alone.
    \item \textbf{WordLen transfers much better here than on BBC News.} It remains surprisingly strong into the middle regime, indicating that conversational text is more tolerant of length-based deletions.
    \item \textbf{Llama SFT is strongest throughout the table.} Fine-tuned Llama leads at all five reported rates and substantially improves over Llama base, with the clearest gains in the middle-to-low retention regime.
    \item \textbf{Character-level targeting is strong only when retention is high.} WordFreq-3 (char.) is excellent at mild compression, but its advantage collapses once the skeleton becomes sparse, where token-level methods recover more gracefully.
\end{enumerate}

\textbf{ROUGE-L and CER.}
Tables~\ref{tab:reddit_rougeL} and \ref{tab:reddit_cer} report ROUGE-L and CER
on the same Reddit test set.

\begin{table*}[tbp]
\centering
\small
\caption{ROUGE-L F1 on Reddit conversational text.
\textbf{Bold} = best per column.}
\label{tab:reddit_rougeL}
\begin{tabular}{lccccc}
\toprule
\textbf{Method} & $\rkeep{=}0.9$ & $\rkeep{=}0.7$ & $\rkeep{=}0.5$ & $\rkeep{=}0.3$ & $\rkeep{=}0.1$ \\
\midrule
Step              & $0.9799$ & $0.8019$ & $0.1614$ & $0.1012$ & $0.0550$ \\
WordLen           & $0.9869$ & $\mathbf{0.9725}$ & $0.6588$ & $0.2142$ & $0.0763$ \\
WordFreq-3         & $0.9411$ & $0.8253$ & $0.6518$ & $0.4392$ & $0.1742$ \\
WordFreq-3 (char.) & $\mathbf{0.9871}$ & $0.9724$ & $0.6619$ & $0.2105$ & $0.0732$ \\
WordFreq-6         & $0.9446$ & $0.8190$ & $0.6372$ & $0.4135$ & $0.1478$ \\
LP            & $0.9821$ & $0.9190$ & $\mathbf{0.7718}$ & $\mathbf{0.5563}$ & $0.1876$ \\
Entropy           & $0.9784$ & $0.9096$ & $0.7382$ & $0.4722$ & $\mathbf{0.1903}$ \\
Entropy-LP        & $0.9789$ & $0.9146$ & $0.7437$ & $0.4754$ & $0.1897$ \\
Entropy-in-FreqBuckets & $0.9844$ & $0.8552$ & $0.6570$ & $0.4341$ & $0.1876$ \\
Hybrid-$0.3$      & $0.9780$ & $0.8977$ & $0.7317$ & $0.4705$ & $0.1846$ \\
Hybrid-$0.5$      & $0.9792$ & $0.8874$ & $0.6886$ & $0.4526$ & $0.1758$ \\
Hybrid-$0.7$      & $0.9774$ & $0.8919$ & $0.6548$ & $0.3827$ & $0.1518$ \\
\bottomrule
\end{tabular}
\end{table*}

\begin{table*}[tbp]
\centering
\small
\caption{Character Error Rate (CER, lower is better) on Reddit conversational text.
\textbf{Bold} = best (lowest) per column.}
\label{tab:reddit_cer}
\begin{tabular}{lccccc}
\toprule
\textbf{Method} & $\rkeep{=}0.9$ & $\rkeep{=}0.7$ & $\rkeep{=}0.5$ & $\rkeep{=}0.3$ & $\rkeep{=}0.1$ \\
\midrule
Step              & $0.0118$ & $0.1329$ & $0.5896$ & $0.7665$ & $0.8946$ \\
WordLen           & $0.0108$ & $\mathbf{0.0183}$ & $0.2349$ & $0.5895$ & $0.8378$ \\
WordFreq-3         & $0.0795$ & $0.2218$ & $0.4129$ & $0.6398$ & $0.8813$ \\
WordFreq-3 (char.) & $\mathbf{0.0106}$ & $\mathbf{0.0183}$ & $\mathbf{0.2337}$ & $0.5959$ & $0.8449$ \\
WordFreq-6         & $0.0654$ & $0.2209$ & $0.4349$ & $0.6915$ & $0.8998$ \\
LP            & $0.0220$ & $0.1271$ & $0.3339$ & $0.5932$ & $0.8842$ \\
Entropy           & $0.0271$ & $0.1111$ & $0.2910$ & $0.5408$ & $0.7830$ \\
Entropy-LP        & $0.0273$ & $0.1072$ & $0.2863$ & $0.5348$ & $\mathbf{0.7747}$ \\
Entropy-in-FreqBuckets & $0.0205$ & $0.2032$ & $0.4411$ & $0.6159$ & $0.7875$ \\
Hybrid-$0.3$      & $0.0262$ & $0.1144$ & $0.2891$ & $\mathbf{0.5339}$ & $0.7787$ \\
Hybrid-$0.5$      & $0.0239$ & $0.1201$ & $0.3204$ & $0.5407$ & $0.8535$ \\
Hybrid-$0.7$      & $0.0251$ & $0.1078$ & $0.3396$ & $0.6028$ & $0.8384$ \\
\bottomrule
\end{tabular}
\end{table*}

These lexical diagnostics sharpen the Reddit picture. ROUGE-L favors methods that
preserve longer contiguous spans, so character-targeted retention and WordLen are
strongest at high retention, while LP becomes strongest in the middle regime.
At the most aggressive setting, Entropy, Entropy-LP, LP and Entropy-in-FreqBuckets
are within $0.003$ ROUGE-L of one another, so no method keeps a clear lexical edge. CER tells a related but slightly different story. At high
retention, character-targeted and length-based methods make the smallest surface
edits. Once compression becomes more aggressive, however, the best CER shifts to
the semantic family: Hybrid-$0.3$ is lowest at $\rkeep = 0.3$, and Entropy-LP is
lowest at $\rkeep = 0.1$. Together, Tables~\ref{tab:reddit_rougeL}
and~\ref{tab:reddit_cer} support the same qualitative conclusion as BERTScore:
conversational text benefits from contextual deletion signals, especially once
the retained skeleton becomes sparse.

\section{Extension to Chinese Text}
\label{app:chinese_eval}

We evaluate three Chinese datasets: Chinese official news,
Chinese Wikipedia, and Zhihu answers. The first two are closer to the English
news and encyclopedic datasets used elsewhere in the paper; Zhihu provides a
more conversational, opinion-heavy register. All three Chinese datasets include semantic/hybrid sweeps (Entropy, Entropy-LP,
Entropy-in-FreqBuckets, and Hybrid-$\alpha$) using the same Gemini 2.0 Flash
decoder, and all three also include accompanying Qwen base/SFT decoder
results using Alibaba's Qwen2.5-3B-Instruct model from the Qwen2.5 family of
multilingual open-weight language models~\citep{yang2024qwen25}.

\textbf{Tokenization.}
Chinese text requires explicit word segmentation since it lacks whitespace boundaries.
We process Chinese text using HanLP~\citep{hanlp}, joining segmented tokens with
a ``/'' delimiter to obtain a whitespace-separated representation suitable for frequency
lookup. Frequency buckets are populated using Simplified Chinese Zipf scores from
\textit{wordfreq}~\citep{speer2022wordfreq}. Surprisal for the Chinese semantic and hybrid rows comes from a Chinese GPT-2, \texttt{uer/gpt2-chinese-cluecorpussmall}~\citep{zhao2019uer}.

\subsection*{Chinese Official News}

We first evaluate on the Chinese Official Daily News corpus~\citep{chinesenews2016kaggle},
using a held-out test set of $n{=}200$ chunks. Table~\ref{tab:chinese_bertscore} reports
BERTScore F1 across the frequency-family, entropy-based, and hybrid deletion strategies
with Gemini 2.0 Flash as the decoder. The LP methods use character-level frequency weights
computed over the segmented tokens. 

\begin{table*}[tbp]
\centering
\small
\caption{BERTScore F1 ($n=200$) on Chinese official news text
for Gemini 2.0 Flash zero-shot reconstruction, with additional local Qwen
base/SFT decoder results.
Entropy and Hybrid rows use the local Chinese GPT-2 surprisal pipeline.
\textbf{Bold} = best per column within each zero-shot / Qwen block.}
\label{tab:chinese_bertscore}
\begin{tabular}{lccccc}
\toprule
\textbf{Method} & $\rkeep{=}0.9$ & $\rkeep{=}0.7$ & $\rkeep{=}0.5$ & $\rkeep{=}0.3$ & $\rkeep{=}0.1$ \\
\midrule
Step              & $0.9843$ & $0.9468$ & $0.8636$ & $0.7389$ & $0.6512$ \\
WordLen       & $0.9836$ & $0.9502$ & $0.8648$ & $0.7351$ & $0.6371$ \\
WordFreq    & $0.9831$ & $\mathbf{0.9514}$ & $\mathbf{0.8703}$ & $0.7384$ & $0.6368$ \\
LP   & $0.9803$ & $0.9382$ & $0.8507$ & $0.7583$ & $0.6626$ \\
Entropy                 & $0.9818$ & $0.9371$ & $0.8498$ & $0.7559$ & $0.6455$ \\
Entropy-LP              & $0.9823$ & $0.9395$ & $0.8496$ & $0.7594$ & $0.6523$ \\
Entropy-in-FreqBuckets  & $\mathbf{0.9910}$ & $0.9235$ & $0.8253$ & $0.7432$ & $0.6271$ \\
Hybrid-$0.3$            & $0.9760$ & $0.9218$ & $0.8481$ & $\mathbf{0.7659}$ & $\mathbf{0.6699}$ \\
Hybrid-$0.5$            & $0.9742$ & $0.9068$ & $0.8337$ & $0.7621$ & $0.6665$ \\
Hybrid-$0.7$            & $0.9747$ & $0.9087$ & $0.8191$ & $0.7394$ & $0.6594$ \\
\midrule
Qwen base              & $0.9502$ & $0.8347$ & $0.6632$ & $0.6831$ & $0.5712$ \\
Qwen SFT               & $\mathbf{0.9828}$ & $\mathbf{0.9353}$ & $\mathbf{0.8652}$ & $\mathbf{0.7451}$ & $\mathbf{0.6610}$ \\
\bottomrule
\end{tabular}
\end{table*}

\textbf{Key findings:}
\begin{enumerate}
    \item \textbf{The overall framework transfers well to Chinese news.} At mild compression, nearly all methods remain very strong, confirming that LLM-assisted reconstruction is not limited to English.
    \item \textbf{Frequency dominates the middle regime.} WordFreq is strongest at $\rkeep=0.7$ and $0.5$, so the main English pattern largely carries over here.
    \item \textbf{Semantic and hybrid methods matter most at aggressive compression.} Once retention becomes very low, Hybrid-$0.3$ and Entropy-LP overtake the frequency-family baselines.
    \item \textbf{Chinese SFT helps, but does not overturn the zero-shot ranking.} Qwen SFT is much stronger than the unfine-tuned local decoder, yet it generally remains below the best Gemini zero-shot deletion setups.
    \item \textbf{The low-retention regime is harder than in English.} The Chinese curves fall further at $\rkeep=0.1$, consistent with higher information density per character.
\end{enumerate}

\subsection*{Chinese Wikipedia}
\label{app:chinese_wiki_eval}

We additionally evaluate on Chinese Wikipedia articles to assess whether the same trends
hold beyond formal news text. The dataset consists of $n{=}200$ held-out chunks
($\le 512$ characters) from the ROOTS Chinese Wikipedia corpus~\citep{roots_zh_wiki}.
The same HanLP tokenization, Zipf frequency buckets, and Gemini 2.0 Flash decoder are used
(Table~\ref{tab:chinese_wiki_bertscore}).

\begin{table*}[tbp]
\centering
\small
\caption{BERTScore F1 ($n=200$) on Chinese Wikipedia
for Gemini 2.0 Flash zero-shot reconstruction, with additional local Qwen
base/SFT decoder results.
Entropy and Hybrid rows use the local Chinese GPT-2 surprisal pipeline.
\textbf{Bold} = best per column within each zero-shot / Qwen block.}
\label{tab:chinese_wiki_bertscore}
\begin{tabular}{lccccc}
\toprule
\textbf{Method} & $\rkeep{=}0.9$ & $\rkeep{=}0.7$ & $\rkeep{=}0.5$ & $\rkeep{=}0.3$ & $\rkeep{=}0.1$ \\
\midrule
Step              & $0.9686$ & $0.9213$ & $0.8303$ & $0.7214$ & $0.6286$ \\
WordLen       & $0.9749$ & $0.9241$ & $\mathbf{0.8430}$ & $0.7236$ & $0.6265$ \\
WordFreq    & $0.9761$ & $0.9242$ & $0.8405$ & $0.7252$ & $0.6268$ \\
LP   & $0.9663$ & $0.9061$ & $0.8211$ & $0.7402$ & $0.6343$ \\
Entropy                 & $0.9754$ & $0.9127$ & $0.8273$ & $0.7543$ & $0.6515$ \\
Entropy-LP              & $0.9766$ & $0.9136$ & $0.8285$ & $\mathbf{0.7560}$ & $\mathbf{0.6563}$ \\
Entropy-in-FreqBuckets  & $\mathbf{0.9849}$ & $\mathbf{0.9307}$ & $0.8352$ & $0.7493$ & $0.6481$ \\
Hybrid-$0.3$            & $0.9630$ & $0.8934$ & $0.8204$ & $0.7518$ & $0.6510$ \\
Hybrid-$0.5$            & $0.9568$ & $0.8758$ & $0.8075$ & $0.7470$ & $0.6413$ \\
Hybrid-$0.7$            & $0.9579$ & $0.8696$ & $0.7850$ & $0.7189$ & $0.6188$ \\
\midrule
Qwen base              & $0.9527$ & $0.8055$ & $0.6638$ & $0.6747$ & $0.5680$ \\
Qwen SFT               & $\mathbf{0.9763}$ & $\mathbf{0.9078}$ & $\mathbf{0.8332}$ & $\mathbf{0.7058}$ & $\mathbf{0.6286}$ \\
\bottomrule
\end{tabular}
\end{table*}

\textbf{Key findings:}
\begin{enumerate}
    \item \textbf{Chinese Wikipedia is harder than Chinese news.} The whole table sits below the news results, indicating a more difficult reconstruction problem.
    \item \textbf{Semantic bucket allocation helps most at mild compression.} Entropy-in-FreqBuckets is strongest at $\rkeep=0.9$ and $0.7$, so Chinese Wikipedia behaves more like an encyclopedic stress test than a news dataset.
    \item \textbf{WordLen remains useful in the middle regime.} At $\rkeep=0.5$, it still edges out the other zero-shot methods.
    \item \textbf{Entropy-LP is the strongest low-retention method.} At $\rkeep=0.3$ and $0.1$, semantic bucket allocation works better than the direct hybrids.
    \item \textbf{Chinese Wikipedia is less favorable to local fine-tuning.} Qwen SFT improves substantially over the base decoder, but it does not surpass the strongest Gemini zero-shot methods.
    \item \textbf{The Chinese news and Chinese Wikipedia stories are meaningfully different.} News favors Hybrid-$0.3$ at the lowest rates, whereas Chinese Wikipedia favors Entropy-LP, suggesting that encyclopedic text benefits more from semantic allocation than from direct frequency--surprisal interpolation.
\end{enumerate}

\subsection*{Zhihu}
\label{app:zhihu_eval}

We further evaluate on Zhihu answers~\citep{wang2024zhihukol}, which differ substantially from the previous
two Chinese datasets in style: they are more conversational, argumentative, and
user-authored, with less formulaic sentence structure than official news and less
topic regularity than Wikipedia. For Zhihu we report both the frequency-family
methods and a separate semantic/hybrid sweep based on Chinese GPT-2 surprisal (Table~\ref{tab:zhihu_bertscore}).

\begin{table*}[tbp]
\centering
\small
\caption{BERTScore F1 ($n=200$) on Zhihu
answers for Gemini 2.0 Flash reconstruction, with additional local Qwen
base/SFT decoder results.  Entropy and Hybrid rows
use the local Chinese GPT-2 surprisal pipeline. \textbf{Bold} = best per column within each zero-shot / Qwen block.}
\label{tab:zhihu_bertscore}
\begin{tabular}{lccccc}
\toprule
\textbf{Method} & $\rkeep{=}0.9$ & $\rkeep{=}0.7$ & $\rkeep{=}0.5$ & $\rkeep{=}0.3$ & $\rkeep{=}0.1$ \\
\midrule
Step          & $0.9622$ & $0.9023$ & $0.7889$ & $0.6873$ & $0.6231$ \\
WordLen   & $0.9692$ & $\mathbf{0.9089}$ & $\mathbf{0.8069}$ & $0.6932$ & $0.6221$ \\
WordFreq& $0.9702$ & $0.9066$ & $0.8000$ & $0.6910$ & $0.6182$ \\
LP  & $0.9561$ & $0.8938$ & $0.7878$ & $0.7033$ & $0.6301$ \\
Entropy                 & $0.9634$ & $0.9055$ & $0.8047$ & $0.7232$ & $0.6248$ \\
Entropy-LP              & $0.9636$ & $0.9046$ & $0.8027$ & $\mathbf{0.7341}$ & $\mathbf{0.6431}$ \\
Entropy-in-FreqBuckets  & $\mathbf{0.9780}$ & $0.8834$ & $0.7732$ & $0.6993$ & $0.6060$ \\
Hybrid-$0.3$            & $0.9630$ & $0.8912$ & $0.8037$ & $0.7310$ & $0.6374$ \\
Hybrid-$0.5$            & $0.9624$ & $0.8799$ & $0.7947$ & $0.7325$ & $0.6427$ \\
Hybrid-$0.7$            & $0.9629$ & $0.8861$ & $0.7874$ & $0.7124$ & $0.6331$ \\
\midrule
Qwen base              & $0.9506$ & $0.8000$ & $0.6469$ & $0.6482$ & $0.5731$ \\
Qwen SFT               & $\mathbf{0.9907}$ & $\mathbf{0.9378}$ & $\mathbf{0.8980}$ & $\mathbf{0.8377}$ & $\mathbf{0.5979}$ \\
\bottomrule
\end{tabular}
\end{table*}

\textbf{Key findings:}
\begin{enumerate}
    \item \textbf{Zhihu has a different best-method profile from the other Chinese datasets.} At very high retention, Entropy-in-FreqBuckets is strongest, while the middle regime looks more favorable to WordLen.
    \item \textbf{Length and frequency heuristics still dominate the middle regime.} The semantic and hybrid methods do not control $\rkeep=0.7$ or $0.5$ here.
    \item \textbf{Entropy-LP is strongest once retention becomes aggressive.} At $\rkeep=0.3$ and $0.1$, semantic LP overtakes both LP and the direct hybrids.
    \item \textbf{Zhihu benefits the most from local fine-tuning.} Qwen SFT is the strongest method from $\rkeep=0.9$ through $0.3$, indicating that this conversational domain is especially favorable to local adaptation.
    \item \textbf{Hybrid gains are present but secondary.} The direct hybrids improve on some low-retention frequency baselines, but none surpass Entropy-LP.
\end{enumerate}

\section{Example Reconstructions}
\label{app:examples}

Tables~\ref{tab:examples} and~\ref{tab:examples_zh} give single illustrative
examples across multiple schemes and retention rates in English and Chinese.
They are included for qualitative intuition only: each row is just one example,
and the displayed BERTScores should not be compared directly to the dataset-level
means in the main tables and figures.

\begin{table*}[tp]
\centering
\small
\caption{Single illustrative English reconstructions from BBC test articles. The displayed compressed strings are generated from the corresponding single-sentence deletion schemes, so the labeled $\rkeep$ values should be read as single-example illustrations rather than dataset-level averages. BS = BERTScore F1 for this single example only.}
\label{tab:examples}
\begin{tabular}{p{1.5cm}p{0.9cm}p{2.7cm}p{2.5cm}p{2.7cm}p{0.85cm}}
\toprule
\textbf{Scheme} & $\rkeep$ & \textbf{Original} & \textbf{Compressed} & \textbf{Recovered} & \textbf{BS} \\
\midrule
Step & 0.9 & He said his party was the one of hope and was ready for a 2005 poll. &
He said hs party ws the on of hope nd was rady for a2005 pol. &
He said his party was the one of hope and was ready for a 2005 poll. & 1.000 \\
\midrule
WordFreq & 0.7 & P2P nets can be used to share any kind of file, like photos, free software, licensed music and any other digital content. &
P2P nets used share kind file, photos, free software, licensed music digital content. &
P2P networks are used to share any kind of file, including photos, free software, licensed music, and other digital content. & 0.979 \\
\midrule
Entropy & 0.5 & Labour and the Liberal Democrats have said the party's sums do not add up and claim it would cut frontline services. &
and Liberal said the sums do add claim would cut frontline &
Labour and the Liberal Democrats said the sums do not add up, and claim would cut frontline services. & 0.963 \\
\midrule
Hybrid-$0.5$ & 0.3 & Swashbuckling adventure film Romancing The Stone saw Douglas's female co-star Kathleen Turner win a Golden Globe award for her performance in 1985. &
Swash adventure Rom Douglas's Kathleen 1985. &
Swashbuckling adventure Romancing the Stone, starring Michael Douglas's Kathleen Turner, was released in 1985. & 0.926 \\
\midrule
Summ. & 0.1 & Paid-for-pass-along, in which firms receive money each time a file is shared, along with various DRM solutions and advertiser-based options are all being considered. &
Pay-per-share, DRM, ads: considered. &
Pay-per-share, DRM, ads: All considered. & 0.884 \\
\bottomrule
\end{tabular}
\end{table*}

\begin{table*}[!t]
\centering
\small
\caption{Single illustrative Chinese reconstructions from a held-out evaluation sample. To keep the table compact, we show only the first-sentence excerpt of the original and the corresponding saved degraded / reconstructed prefixes for each method; the displayed compressed strings remain close to the labeled $\rkeep$ values. BS = BERTScore F1 for this single example only.}
\label{tab:examples_zh}
\begin{CJK*}{UTF8}{gbsn}
\begin{tabular}{p{1.6cm}p{0.9cm}p{2.7cm}p{2.5cm}p{2.7cm}p{0.85cm}}
\toprule
\textbf{Scheme} & $\rkeep$ & \textbf{Original} & \textbf{Compressed} & \textbf{Recovered} & \textbf{BS} \\
\midrule
Entropy & 0.9 & \multirow{5}{2.7cm}{中国和澳大利亚外长今天(17日)在北京举行第三轮中澳外交与战略对话后，共同会见记者，在回答关于如何破解当前朝鲜半岛核问题困局时，中方提出实现半岛无核化与停和机制转换并行推进的谈判思路。} & 中国和澳大利亚外长今天(17日在北京举行第三轮中澳外交与战略对话后共同会见记者，在回答关于如何破解当前朝鲜半岛核问题困局时中方提出实现半岛无核化与停和机制转换并行推进 & 中国和澳大利亚外长今天(17日)在北京举行第三轮中澳外交与战略对话后共同会见记者，在回答关于如何破解当前朝鲜半岛核问题困局时，中方提出实现半岛无核化与停和机制转换并行推进的谈判思路。 & 0.996 \\
\cmidrule(lr){1-2}\cmidrule(lr){4-6}
Hybrid-$0.7$ & 0.7 &  & 澳大利亚外长今天()第三轮澳外交后，共同会见记者，回答关于如何破解当前朝鲜半岛核困局，中方实现半岛无核化停和机制转换并行推进谈判思路。 & 澳大利亚外长今天结束第三轮澳中外交与战略对话后，共同会见记者，回答了关于如何破解当前朝鲜半岛核困局，以及中方关于实现半岛无核化与停和机制转换并行推进的谈判思路。 & 0.909 \\
\cmidrule(lr){1-2}\cmidrule(lr){4-6}
Ent.-FreqBkt. & 0.5 &  & 中国和今天17日在北京举行中与战略对话后共同记者在回答关于如何问题时提出实现与停和的停和的也 & 中国和巴西今天17日在北京举行中巴高级别战略对话后共同会见记者。 & 0.739 \\
\cmidrule(lr){1-2}\cmidrule(lr){4-6}
Hybrid-$0.3$ & 0.3 &  & 和外长今天(第三轮外交与后破解困局中方实现无核化与停和机 & 中国外交部长今天表示，在第三轮外交与安全对话中，破解半岛困局的关键在于实现半岛无核化与建立半岛和平机制并行推进。 & 0.769 \\
\cmidrule(lr){1-2}\cmidrule(lr){4-6}
Hybrid-$0.5$ & 0.1 &  & 外长(第三轮外交困 & 王毅外长在第三轮外交斡旋中强调，当前外交面临困局，各方应并行推进停战止暴，为东北亚地区的长治久安创造前景。 & 0.665 \\
\bottomrule
\end{tabular}
\end{CJK*}
\end{table*}

\section{Prompt Templates}
\label{app:prompts}

\textbf{Reconstruction prompt, semantic and hybrid strategies (English):}

\begin{quote}
\textit{You are a text reconstruction assistant. The input
is a \textup{[domain]} from which some tokens have been deleted.
Reconstruct the original text as faithfully as possible.
Do not add new facts, names, or events not suggested by the
input. The output should be approximately the same length
(\textasciitilde\textup{[n]} characters). Return only the
reconstructed text.}
\end{quote}
\noindent Here \textup{[domain]} is ``BBC news article excerpt'', ``Wikipedia article excerpt'' or
``Reddit comment'', and \textup{[n]} is the length of the original in characters.

\textbf{Reconstruction prompt, semantic and hybrid strategies (Chinese):}
\begin{quote}
\begin{CJK*}{UTF8}{gbsn}
\textit{你是一个中文文本重建助手。输入是一段\textup{[domain]}，
其中部分词语已被删除。请在不添加输入中没有依据的新事实、
新人物或新事件的前提下，尽可能忠实地重建原文。输出长度应
大致接近原文（约\textup{[n]}个字符）。只返回重建后的中文文本，
不要加解释。}
\end{CJK*}
\end{quote}
\noindent Here \textup{[domain]} is the English label ``Chinese news article'', ``Chinese Wikipedia article excerpt''
or ``Zhihu answer''.

\textbf{Reconstruction prompt, word-level strategies (English):}
\begin{quote}
\textit{You are a linguistic reconstruction model whose goal is to recover the original text from corrupted
input. The text may have missing English words (due to category-based deletion), missing punctuation
symbols ({\em e.g.}, commas, periods, hyphens), missing numeric or non-alphabetic tokens, or missing whitespace
characters such as spaces, line breaks, or tabs. Restore the text to its closest original English form. Do
NOT paraphrase. Do NOT add new meaning. Only restore missing parts.}
\end{quote}
The BBC Step reconstructions reported with the word-level strategies ({\em e.g.}, Table~\ref{tab:rougel}) use the
same prompt with the second sentence describing missing letters, short words, or partially deleted characters.

\textbf{Reconstruction prompt, character-level strategies (English):}
\begin{quote}
\textit{You are a text reconstruction assistant. The input is an English news article from which some
characters have been deleted. Your task is to reconstruct the original article as faithfully as possible.
Do not add new facts, names, or events not suggested by the input. The output should be approximately
\textup{[n]} characters long. Return only the reconstructed text.}
\end{quote}

\textbf{Summarization baseline of Table~\ref{tab:semantic_baselines} (English):}
\begin{quote}
\textit{Compress the following news text into exactly \textup{[m]} characters ($\pm$15\%). Preserve all key
facts, names, numbers, and events. Do not add any new information. Return only the compressed text.}
\end{quote}
\noindent Here \textup{[m]} is $\rkeep$ times the length of the original.

\textbf{Instruction for Llama-3.2-3B-Instruct, zero-shot and fine-tuned (English):}
\begin{quote}
\textit{The following text has been corrupted \textup{[deletion]} (keep ratio: \textup{[r]}). Reconstruct the
original text faithfully, preserving the original wording and meaning. Do not paraphrase or introduce new
content. Return only the recovered text.}
\end{quote}
\noindent For Table~\ref{tab:finetune}, \textup{[deletion]} is ``by character, punctuation, number, and whitespace
deletion'' for LP ($\rkeep = 0.9, 0.5$), ``through structured, ratio-controlled deletions involving partial space
removal and progressive letter reduction within words'' for WordLen ($0.7$), and ``by proportionally deleting words
based on low-, mid-, and high-frequency categories'' for WordFreq-3 ($0.3, 0.1$). The Wikipedia and Reddit runs
use the same instruction with minor wording differences.

\textbf{Instruction for Qwen2.5-3B-Instruct, zero-shot and fine-tuned (Chinese):}
\begin{quote}
\begin{CJK*}{UTF8}{gbsn}
\textit{根据以下残缺文本骨架，补全缺失的汉字、标点和数字，还原为完整、通顺的原文。}\\
\textit{要求：}\\
\textit{1. 骨架中的所有字符必须原样保留，顺序不能改变}\\
\textit{2. 只能在字符之间插入内容，不能删除或替换已有字符}\\
\textit{3. 不得添加原文中不存在的信息}\\
\textit{4. 输出应自然连贯，符合正常语言表达}\\
\textit{5. 只输出最终还原文本，不要输出任何解释}\\
\textit{残留文本保留比例：\textup{[r]}}
\end{CJK*}
\end{quote}
\noindent Two fine-tuned runs differ: Chinese news at $\rkeep = 0.5$ uses a longer instruction with the same
constraints, and Chinese Wikipedia and Zhihu at $\rkeep = 0.1$ append a sentence asking to restore the missing
$1 - \rkeep$ of the text. The instruction used for every test item is stored with the released generations.

\section{LP-Guided Allocation: Extended Derivation}
\label{app:lp_derivation}

We provide additional derivation of the objective in Section~\ref{sec:optimize} and of the linear program
that gives the method its name.

\textbf{Rationale for linear interpolation.}
The linear model $B_k(w_k) = 1 - w_k(1 - B_k^{\mathrm{full}})$
follows from two boundary conditions: (i) when nothing
is deleted from bucket $k$ ($w_k = 0$), that bucket
contributes perfectly to the BERTScore, so $B_k(0) = 1$;
(ii) when the entire bucket is deleted ($w_k = 1$), the
contribution falls to $B_k^{\mathrm{full}}$, measured empirically.
The linear interpolation between these two points is the
simplest assumption.

\textbf{Independence assumption.}
The product objective $\prod_k B_k(w_k)$ assumes that the
score losses caused by different buckets compound
independently: $B_k^{\mathrm{full}}$ is measured on the whole
text with bucket $k$ removed, so it is a retention factor for
the document, and the factors are multiplied.  In reality,
deleting all content from two buckets simultaneously may
cause worse recovery than their factors predict.  This is a
limitation of the first-order model that future work should
address with interaction terms.

\textbf{The linear program.}
Writing $a_k = 1 - B_k^{\mathrm{full}}$, the first-order expansion
$\prod_k (1 - a_k w_k) \approx 1 - \sum_k a_k w_k$ turns the
problem into a bounded-variable LP.  Let
$\mathbf{w} = [w_1, \ldots, w_K]^\top$:
\begin{equation*}
  \min_{\mathbf{w}} \ \mathbf{a}^\top \mathbf{w}
  \quad \text{s.t.} \quad \mathbf{p}^\top \mathbf{w} \ge 1 - \rkeep,\ \ \mathbf{0} \le \mathbf{w} \le \mathbf{1},
\end{equation*}
where $p_k$ is the character fraction of bucket $k$.  Its
closed form sorts buckets in ascending order of $a_k / p_k$,
the score lost per unit of deletion budget, and greedily sets
$w_k = 1$ until the budget is exhausted.  Because $a_k$ already
reflects the size of bucket $k$, the objective carries no $p_k$ weight. Weighting it by $p_k$ would count size twice and
would delete rare words first.  On the $200$ BBC test articles
at nine retention rates this LP reproduces the product's
allocation in $96\%$ of cells (no $w_k$ differs by more than
$0.02$).  The hybrids Entropy-LP and Entropy-in-FreqBuckets
(Section~\ref{sec:hybrid}) use the additive variant
$\sum_k p_k B_k(w_k)$, solved greedily in ascending order of
$a_k$, with $B_k^{\mathrm{full}}$ calibrated once on the first $50$ of the $100$ test chunks.

\textbf{Hybrid implementation details.}
For the semantic and hybrid strategies the unit of deletion is the GPT-2 token, including its leading
space, and the budget is $n_{\mathrm{keep}} = \max(1, \lfloor \rkeep \cdot n \rfloor)$ tokens, so the realised
character ratio differs from $\rkeep$: on the $100$ BBC chunks of Table~\ref{tab:semantic_baselines} Entropy,
Entropy-LP and Hybrid-$\alpha{=}0.5$ keep $.92$/$.75$--$.77$/$.56$--$.59$/$.36$--$.38$/$.11$--$.13$ of the
characters at the five nominal rates, against $.90$/$.71$/$.51$/$.32$/$.13$ for WordFreq-3, so at
$\rkeep = 0.7$--$0.3$ the token-level encoders keep $0.04$--$0.07$ more text. Entropy keeps the $n_{\mathrm{keep}}$ tokens of highest
GPT-2 surprisal. Hybrid-$\alpha$ scores each token as
$\alpha\,\mathrm{rank}(\text{Zipf}) + (1-\alpha)\,\mathrm{rank}(1/\text{surprisal})$, where
$\mathrm{rank}(\cdot)$ is the stable ascending rank divided by $n-1$, and keeps the $n_{\mathrm{keep}}$
lowest-scoring tokens, {\em i.e.}, the rarest and most surprising ones. Ties are broken by position. Kept tokens
are concatenated in their original order.

\section{Compute Resources}
\label{app:compute}

Fine-tuning experiments were conducted on a single NVIDIA
A100 80~GB GPU.  Each fine-tuning run (3 epochs, about $2{,}000$ training examples, max length 2048 tokens) required
approximately $2$--$4$ hours per compression setting.
Inference with the fine-tuned model runs at approximately
$100$ tokens per second at 4-bit precision.

Gemini 2.0 Flash API calls average $1$--$3$ seconds per
article chunk, including retry logic.  The full Gemini-based
evaluation over the reported methods and retention rates
requires on the order of a few thousand API calls and
several hours of wall-clock time.

\section{Reproducibility Statement}

\label{app:reproducibility}

The paper uses fixed public datasets, explicit retention rates, stored result files, and deterministic post-hoc evaluation scripts where possible. The appendix
reports decoder prompts, compute details, and cross-domain dataset sources. All reported tables can be reproduced from the scripts and stored result files, which we release at \url{https://github.com/junli-cuny/lossy-text-compression-benchmark}.
The release includes the source texts of the public corpora in the form our experiments read them, processed evaluation splits, compressed text
skeletons, reconstructed outputs, and evaluation scripts needed to reproduce the reported
results, subject to the licenses and redistribution terms of the underlying datasets and
models.

\paragraph{Artifacts, terms, and implementation details.}
All datasets, models, and software used in this study are existing public research artifacts cited in the paper, including public benchmark datasets, released open-weight models, hosted API models, and standard software packages. We use these artifacts only for research evaluation and experimental comparison, which is consistent with their intended research use, and the released copies of the source texts are provided for research reproduction only. The BBC News articles originate from the corpus of \citet{greene2006practical}. Because the datasets are public corpora rather than newly collected human-subject data, they may still contain names of public figures, user-written text, or other naturally occurring sensitive or offensive content. We therefore report aggregate results, keep qualitative examples limited, and do not claim suitability for privacy-sensitive deployment settings. For preprocessing and evaluation, English word tokenization follows standard whitespace-based token boundaries; Chinese segmentation uses HanLP~\citep{hanlp}; frequency scores come from \textit{wordfreq}~\citep{speer2022wordfreq}; named-entity preservation uses spaCy NER~\citep{honnibal2020spacy}; optimization uses CVXPY~\citep{diamond2016cvxpy}; and reconstruction quality is measured with BERTScore~\citep{zhang2020bertscore}, ROUGE-L~\citep{lin2004rouge}, and CER using the model backbones and settings stated in the paper or package defaults unless otherwise noted. chrF++ uses sacreBLEU~\citep{post2018call}, METEOR uses NLTK~\citep{bird2009natural}, Cohen's $\kappa$ uses scikit-learn~\citep{pedregosa2011scikit}, and the Wilcoxon and Spearman tests use SciPy~\citep{virtanen2020scipy}. Fine-tuning trains LoRA adapters~\citep{hu2022lora} on the 4-bit base model with Transformers~\citep{wolf2020transformers}, PEFT~\citep{mangrulkar2022peft}, and TRL~\citep{vonwerra2020trl}.

\section{Related Work}
\label{app:related_work}

\textbf{Traditional Lossless Text Compression.}
Classical algorithms such as Lempel-Ziv~(zlib~\citep{zlib}),
Burrows-Wheeler~(bzip2~\citep{bz2}), and
LZMA~\citep{lzma} achieve lossless compression of natural
language by exploiting statistical regularities and repetition.
For English news text, these algorithms typically
provide only modest compression on practical text segments (and
at the 512-character chunk level used in our experiments, measured
ratios are ${\sim}1.7$--$1.9\times$; see Appendix~\ref{app:lossless_baselines}). Their primary limitation
is that they operate on byte sequences without semantic
understanding, placing a hard ceiling on compression ratios
achievable with lossless reconstruction. In our setting, they are best viewed as complementary to semantic deletion, since they can be applied again to the retained skeleton in a cascaded pipeline.

\textbf{Language Models as Compressors.}
A formal connection between predictive models and lossless
compression has long been established. \citet{deletang2024language} demonstrate
that large pretrained LLMs are near-optimal compressors for
text, outperforming domain-specific codecs on many modalities.
LLMZip~\citep{valmeekam2023llmzip} further applies this idea
directly to \emph{lossless} text compression using arithmetic coding
conditioned on LLM probabilities, and therefore guarantees exact
recovery. That guarantee is also why it admits no numerical comparison
here: a lossless coder has no rate--distortion curve to place beside ours,
since it either reproduces the source exactly or not at all, and its output
is a bitstream rather than readable text. Our work is complementary: we
target the \emph{lossy} regime where full recovery is not required and the
goal is maximizing a semantic fidelity metric at a given compression rate.
The corresponding lossless reference points for our rates are the classical
codecs of Appendix~\ref{app:lossless_baselines}.

\textbf{Neural and Learned Compression.}
Learned compression of images~\citep{minnen2018joint,mandt2019deep}
has demonstrated state-of-the-art performance by learning
end-to-end codecs using rate-distortion objectives.  For text,
analogous systems are less developed because differentiating
through discrete tokens is harder.  Our approach sidesteps the
end-to-end training problem by simply using deletion as the
compression scheme and a pretrained LLM as the decoder. Continuous
semantic bottlenecks are an important complementary direction, but
they assume a different interface from the text-preserving setting
studied here.

\textbf{Semantic Communication.}
Semantic communication systems~\citep{xie2021deep,salehi2025llm}
aim to transmit the \emph{meaning} of a message at minimal
bandwidth.  The DeepSC system~\citep{xie2021deep} trains a
joint source-channel encoder for English sentences, achieving
near-lossless semantic recovery even at low signal-to-noise ratio (SNR).  LLM-enabled
semantic communication~\citep{salehi2025llm} extends
this idea to generation-based receivers.  Our framework shares
this philosophy but focuses on the storage/compression
application rather than the channel-coding setting,
and targets a computationally cheaper regime where only a
deletion mask needs to be transmitted or stored.

\textbf{Prompt Compression and Token Pruning.}
A closely related line of work uses language model perplexity
to decide which tokens to drop from LLM prompts.
LLMLingua~\citep{jiang2023llmlingua} computes per-token
perplexity with a small LM and removes the most predictable
tokens. Selective Context~\citep{li2023compressing} similarly
uses self-information to prune low-information lexical units, and
SCOPE~\citep{zhang2025scope} rewrites chunks generatively instead of
deleting. These methods target \emph{downstream task accuracy} of the
compressed prompt rather than faithful reconstruction of the original text. Selective Context does report a reconstruction task, but as a
diagnostic of what its filter retains, and it is the task that degrades
fastest with compression. Compressor--decompressor frameworks such as
\citet{li2026tokencompressor} run an adapted LLM at both ends and
compress into learned token sequences. Our benchmark fixes the
opposite envelope, a sender with no model weights and a plain-text
artifact, so it answers a complementary question. We compare our frequency-based deletion against a
perplexity-based dropping baseline (Section~\ref{sec:new_baselines}), and we
additionally run LLMLingua-2~\citep{pan2024llmlingua}, the classifier-based successor of~\citet{jiang2023llmlingua}, as a drop-in encoder inside our own
pipeline, so the change of objective is measured rather than asserted
(Appendix~\ref{app:llmlingua2}).  Entropy-based dropping
is strongest at mild-to-moderate compression, but our
WordFreq baseline requires no neural computation at the encoder
({\em i.e.}, only a frequency table lookup), making it suitable for
resource-constrained settings. More importantly, the benchmark
objective here is different: we study later reconstruction of the
source text for storage/transmission/recovery, not only immediate task performance from the compressed prompt.

\textbf{Deletion-Based Sentence Compression and Autoencoding.}
Deleting tokens to shorten text has a long history in sentence
compression, where \citet{filippova2015sentence} train an LSTM to
decide which words to drop while keeping the sentence grammatical. The target is a shorter sentence, not later recovery of the original. Closer
to our setting, SEQ$^3$~\citep{baziotis2019seq3} chains two
encoder--decoder pairs so that a compressed \emph{text} is the latent
representation from which the input is reconstructed, at sentence level
and with a trained, abstractive compressor. Latent alternatives such as
the in-context autoencoder~\citep{ge2024icae} compress a context into
memory slots that only the same model can read. Our benchmark differs
from all three in what is held fixed: document-length inputs, a
training-free deletion encoder, a textual artifact, and reconstruction
by a general pretrained decoder rather than a jointly trained one.

\textbf{Text Summarization as Lossy Compression.}
Abstractive summarization~\citep{rush2015neural,see2017get,liu2019text}
can be viewed as extremely high-ratio lossy compression; at
$5$--$20\times$ compression typical extractive and abstractive
systems discard much of the fine-grained content of the
source.  In contrast, our framework operates at moderate
skeleton retention ($\rkeep = 0.9$ down to $0.1$) 
where near-original
fidelity is required, and uses a deletion-based encoder that
does not require any training of an encoder model.
We include a direct comparison with LLM-based
length-constrained summarization in
Section~\ref{sec:new_baselines}.

\end{document}